\documentclass[11pt]{article}

\usepackage[preprint]{acl}

\usepackage{times}
\usepackage{latexsym}

\usepackage[T1]{fontenc}

\usepackage[utf8]{inputenc}

\usepackage{microtype}

\usepackage{inconsolata}

\usepackage{graphicx}

\usepackage{CJKutf8}
\newcommand{\zh}[1]{\begin{CJK*}{UTF8}{gkai}#1\end{CJK*}}

\usepackage{xcolor}
\newcommand{\err}[1]{\textcolor{red}{#1}}
\newcommand{\cor}[1]{\textcolor{green}{#1}}
\usepackage{tabularx}
\usepackage{booktabs}
\usepackage{multirow}
\usepackage{enumitem}
\usepackage{listings}
\usepackage{hyperref} 
\usepackage{setspace} 
\usepackage{amsmath}
\usepackage{graphicx} 
\newenvironment{tightitemize}{%
  \begin{itemize}[leftmargin=1.2em,itemsep=0.2em,topsep=0.2em,parsep=0em,partopsep=0em]
}{\end{itemize}}

\newcommand{\field}[2]{%
  \noindent\textbf{#1}\hspace{0.6em}#2\par
}
\usepackage[most]{tcolorbox}
\usepackage{fvextra} 

\newtcolorbox{promptbox}[1]{%
  enhanced,
  breakable,
  colback=white,
  colframe=gray!65,
  boxrule=0.5pt,
  arc=1.5mm,
  left=2.2mm,right=2.2mm,top=1.4mm,bottom=1.4mm,
  colbacktitle=gray!65,
  coltitle=white,
  fonttitle=\bfseries\footnotesize,
  title={#1},
  attach boxed title to top left={yshift=-0.5mm, xshift=0.6mm},
  boxed title style={boxrule=0pt, arc=1.5mm, left=1.2mm, right=1.2mm, top=0.6mm, bottom=0.6mm},
  before skip=0.5em,
  after skip=0.5em
}
\newcommand{\PromptRawStart}{%
  \par\begingroup
  \ttfamily\footnotesize
  \setlength{\parindent}{0pt}%
  \setlength{\parskip}{0pt}%
  \raggedright
  \sloppy
  \emergencystretch=1em
  \obeylines\obeyspaces
  \catcode`\#=12 \catcode`\$=12 \catcode`\%=12 \catcode`\&=12
  \catcode`\_=12 \catcode`\~=12 \catcode`\^=12
  \catcode`\{=12 \catcode`\}=12
}

\newcommand{\AgentPromptRawStart}{%
  \par\begingroup
  \ttfamily\fontsize{7.4}{8.1}\selectfont
  \setlength{\parindent}{0pt}%
  \setlength{\parskip}{0pt}%
  \raggedright
  \sloppy
  \emergencystretch=1.2em
  \obeylines\obeyspaces
  \catcode`\#=12 \catcode`\$=12 \catcode`\%=12 \catcode`\&=12
  \catcode`\_=12 \catcode`\~=12 \catcode`\^=12
  \catcode`\{=12 \catcode`\}=12
}

\newcommand{\PromptRawEnd}{\endgroup\par}

\usepackage{xparse}
\NewDocumentEnvironment{promptfigure}{O{t} m m m}
{%
  \begin{figure*}[#1]
  \centering
  \begin{promptbox}{#2}
}
{%
  \end{promptbox}
  \caption{#3}
  \label{#4}
  \end{figure*}
}

\title{CLFEC: A New Task for Unified Linguistic and Factual Error Correction in paragraph-level Chinese Professional Writing}

\author{
\textbf{Jian Kai\textsuperscript{1}}, 
\textbf{Zidong Zhang\textsuperscript{2}}, 
\textbf{Jiwen Chen\textsuperscript{2}}, 
\textbf{Zhengxiang Wu\textsuperscript{2}}, 
\textbf{Songtao Sun\textsuperscript{2}}, 
\textbf{Fuyang Li\textsuperscript{2}} \\
\textbf{Yang Cao\textsuperscript{1}}, 
\textbf{Qiang Liu\textsuperscript{2}} \\
\textsuperscript{1}Huazhong University of Science and Technology \quad
\textsuperscript{2}WPS AI, Kingsoft Office \\
\texttt{\{kaji, ycao\}@hust.edu.cn} \\
\texttt{\{zhangzidong, chenjiwen, wuzhengxiang, sunsongtao, lifuyang, liuqiang2\}@wps.cn}
}

\begin{document}
\maketitle
\begin{abstract}
Chinese text correction has traditionally focused on spelling and grammar, while factual error correction is usually treated separately. However, in paragraph-level Chinese professional writing, linguistic (word/grammar/punctuation) and factual errors frequently co-occur and interact, while many draft-level errors are sparsely observable in published texts after editorial review, making unified correction both necessary and controlled benchmark construction essential. This paper introduces \textbf{CLFEC (Chinese Linguistic \& Factual Error Correction)}, a new task for joint linguistic and factual correction. We construct a mixed, multi-domain Chinese professional writing dataset spanning current affairs, finance, law, and medicine.\footnote{The dataset will be released after internal review.} We then conduct a systematic study of LLM-based correction paradigms, from prompting to retrieval-augmented generation (RAG) and agentic workflows. The analysis reveals practical challenges, including limited generalization of specialized correction models, the need for evidence grounding for factual repair, the difficulty of mixed-error paragraphs, and over-correction on clean inputs. Results further show that handling linguistic and factual errors within the same context outperforms decoupled pipelines, and that agentic workflows can be effective with suitable backbone models. Overall, CLFEC provides a new benchmark for Chinese text correction research and practical guidance for proofreading systems.
\end{abstract}

\section{Introduction}
Research on Chinese text correction has mainly focused on \emph{grammatical error correction} (GEC) and \emph{Chinese spelling correction} (CSC), with the primary goal of improving linguistic well-formedness and standard usage. With continuously updated datasets and advances in modeling, substantial progress has been reported~\cite{wu-etal-2023-rethinking, zhang-etal-2023-nasgec, zhou-etal-2025-training}. Some work further extends sentence-level correction to cross-sentence settings, presenting even greater challenges~\cite{wang-etal-2022-cctc}.

However, inspecting existing resources reveals that some ``gold'' corrections still contain factual or logical errors, which can bias model training. For example, ECSpell~\cite{lv2023general} corrects
\zh{“美国宪法是柔性线法的代表”}
(``The U.S. Constitution is a representative of a flexible \emph{line law}'')
to
\zh{“美国宪法是柔性宪法的代表”}
(``The U.S. Constitution is a representative of a flexible \emph{constitution}''),
fixing the character error
``\zh{线法}$\rightarrow$\zh{宪法}''
(\emph{line law} $\rightarrow$ \emph{constitution})
but leaving a misleading claim, as the U.S.\ Constitution is commonly regarded as \emph{rigid}.
 In contrast, \emph{factual error correction} (FEC) is typically studied separately, focusing on claim extraction, evidence retrieval, and evidence-grounded rewriting~\cite{thorne-vlachos-2021-evidence}.

\begin{table}[t]
\centering
\small
\setlength{\tabcolsep}{3pt}
\renewcommand{\arraystretch}{1.08}
\begin{tabular}{@{}p{0.25\linewidth} p{0.71\linewidth}@{}}
\toprule
\textbf{Error type} & \textbf{Example snippet} \\
\midrule
\textbf{Word} &
\zh{\err{反溃}控制} $\rightarrow$ \zh{反馈控制} \newline
\err{collapse} control $\rightarrow$ feedback control \\
\textbf{Grammatical} &
\zh{\err{增强}...水平} $\rightarrow$ \zh{提高...水平}\newline
\err{Enhance}...level $\rightarrow$ Raise...level \\
\textbf{Punctuation} &
\zh{如下\err{；}} $\rightarrow$ \zh{如下：}\newline
As follows\err{;} $\rightarrow$ As follows: \\
\midrule
\textbf{Factual} &
\zh{\err{六中}全会} $\rightarrow$ \zh{四中全会} \newline
The \err{Sixth} Plenary Session $\rightarrow$ The Fourth Plenary Session \\
\bottomrule
\end{tabular}
\caption{Unifying linguistic and factual error correction, where linguistic errors cover word, grammatical, and punctuation types.}
\vspace{-14pt}
\label{tab:error_types_examples}

\end{table}

Similar issues also appear in our online user feedback data, especially in paragraph-level professional writing. Such texts often mix linguistic issues with non-trivial factual errors, including incorrect entity/event names, outdated regulations, domain-terminology misuse, and inconsistent numerical indicators. When proofreading such texts, human editors bring domain-specific prior knowledge and consult external resources as needed to verify uncertain linguistic usages or factual claims. Meanwhile, knowledge- or semantics-aware augmentation has been shown to improve grammatical and spelling correction~\cite{dong-etal-2024-rich, li-etal-2025-rethinking}, suggesting shared mechanisms between linguistic and factual correction.

We first conduct and release a real corpus analysis by applying three commercial proofreading products to 4 million characters of professional and user-generated content (UGC) texts and manually verifying their correction candidates. The results reveal a mismatch between real-world proofreading needs and direct benchmark construction: many errors in professional texts are no longer observable after editorial review, while UGC errors are more abundant but informal and lack authoritative references.

We introduce \textbf{CLFEC (Chinese Linguistic \& Factual Error Correction)}, a new paragraph-level task for Chinese professional writing that unifies linguistic error correction (LEC) and factual error correction (FEC) in a single setting. Furthermore, we extend linguistic errors into three types: (i) word errors (e.g., including spelling errors and missing/redundant words), (ii) grammatical errors (e.g., non-standard structure, improper collocation, word order problems), and (iii) punctuation errors (Table~\ref{tab:error_types_examples}). A multi-domain dataset is constructed across \textbf{current affairs}, \textbf{finance}, \textbf{law}, and \textbf{medicine}, with four diagnostic splits: LEC, FEC, MIX, and Error-free.

We study CLFEC and characterize the challenges introduced by this task. First, specialized error correction models often exhibit limited generalization under the unseen LEC task. Second, factual errors tend to be specific, time-sensitive, and hard to spot, so reliable correction usually depends on explicit evidence rather than parametric knowledge alone. Third, the MIX task is more challenging, especially when language errors and factual errors overlap, as models often overlook deeper errors. Finally, our fine-grained analysis shows that grammatical and punctuation errors are harder to recall than factual or word errors, likely because these errors are ``weakly normative'' and are often missed even by native speakers.

We then evaluate LLM-based proofreading paradigms, including prompting, retrieval-augmented generation (RAG)~\cite{gao2024retrievalaugmentedgenerationlargelanguage}, and an agentic workflow. We examine over-correction in LLMs~\cite{fang2023chatgpthighlyfluentgrammatical} and find that precision drops as the input becomes cleaner. RAG variants show that handling LEC+FEC in a shared context performs better than decoupled pipelines. Finally, the agentic workflow offers gains for advanced backbones with stronger reasoning ability.

Overall, our work introduces CLFEC, a paragraph-level unified linguistic and factual correction task; releases a real corpus analysis of commercial proofreading products; constructs a multi-domain benchmark via controlled error injection on verified professional texts with LEC, FEC, MIX, and Error-free splits; and evaluates prompting, RAG-based, and agentic workflows to inform reliable proofreading system design.

\section{Real Corpus Analysis}
\label{sec:motivation}
\begin{figure}[b]
\centering
\includegraphics[width=0.9\linewidth]{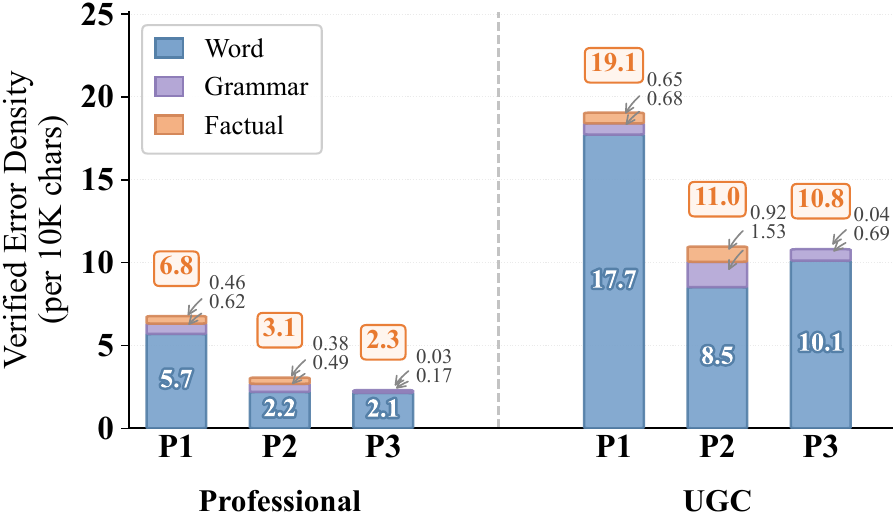}
\caption{Verified error density (per 10K characters) by source type and proofreading product. Professional texts yield few confirmed errors overall, while UGC texts show higher density.}
\label{fig:motivation}
\vspace{-4pt}
\end{figure}

Before constructing CLFEC, we first examine whether a benchmark for Chinese professional proofreading can be directly mined from naturally occurring errors. We conduct an empirical study using three commercial proofreading products (Jinshan, Fangcun, and Fangzheng)\footnote{We anonymize the result-to-product mapping and report them as Product P1, Product P2, and Product P3.} applied to a corpus of 4M characters spanning two source types: Professional texts, primarily from government portal websites (2M characters), and UGC texts, primarily from online forums such as Zhihu (2M characters). We release the full set of product outputs, including all detected candidates, suggested corrections, and our normalized annotations.

We design a mapping scheme to normalize heterogeneous product-side labels into unified error categories, word, grammar, and factual, for aggregate analysis. We then deduplicate overlapping candidates within each product and filter out punctuation- and formatting-related items, which are often stylistic conventions rather than clear-cut errors in this setting. Our annotation team manually verifies the remaining 16K+ candidates. Each candidate is labeled as \textit{false positive}, \textit{confirmed}, \textit{partially valid}, or \textit{uncertain}, where \textit{partially valid} means that the flagged span is problematic but the suggested correction is inappropriate or incomplete.

Figure~\ref{fig:motivation} presents the verified error density (\textit{confirmed}) broken down by source type, product, and error category.
The results show that the available confirmed errors differ sharply between text types.
Professional texts, having undergone editorial review before publication, yield an average of only \textbf{4.06} verified errors per 10K characters across the three products---too sparse to serve as a practical source for benchmark construction.
UGC texts exhibit substantially higher density (\textbf{13.62} per 10K characters on average), but they are predominantly informal and lack authoritative references for ground-truth verification. In addition, the verified candidates from existing products are concentrated on word-level corrections; their factual detections mostly come from rule- or lexicon-based normalization of names, terms, administrative divisions, and units.

These observations motivate our controlled construction strategy. Neither source can directly supply a balanced benchmark for professional-writing correction. Instead of harvesting existing real-world errors, we collect verified clean professional texts and inject errors under controlled settings. This allows CLFEC to preserve professional style and evidence availability while ensuring sufficient density and balanced coverage across linguistic and factual error types.

\begin{figure*}[t]
  \centering
  \includegraphics[width=\textwidth]{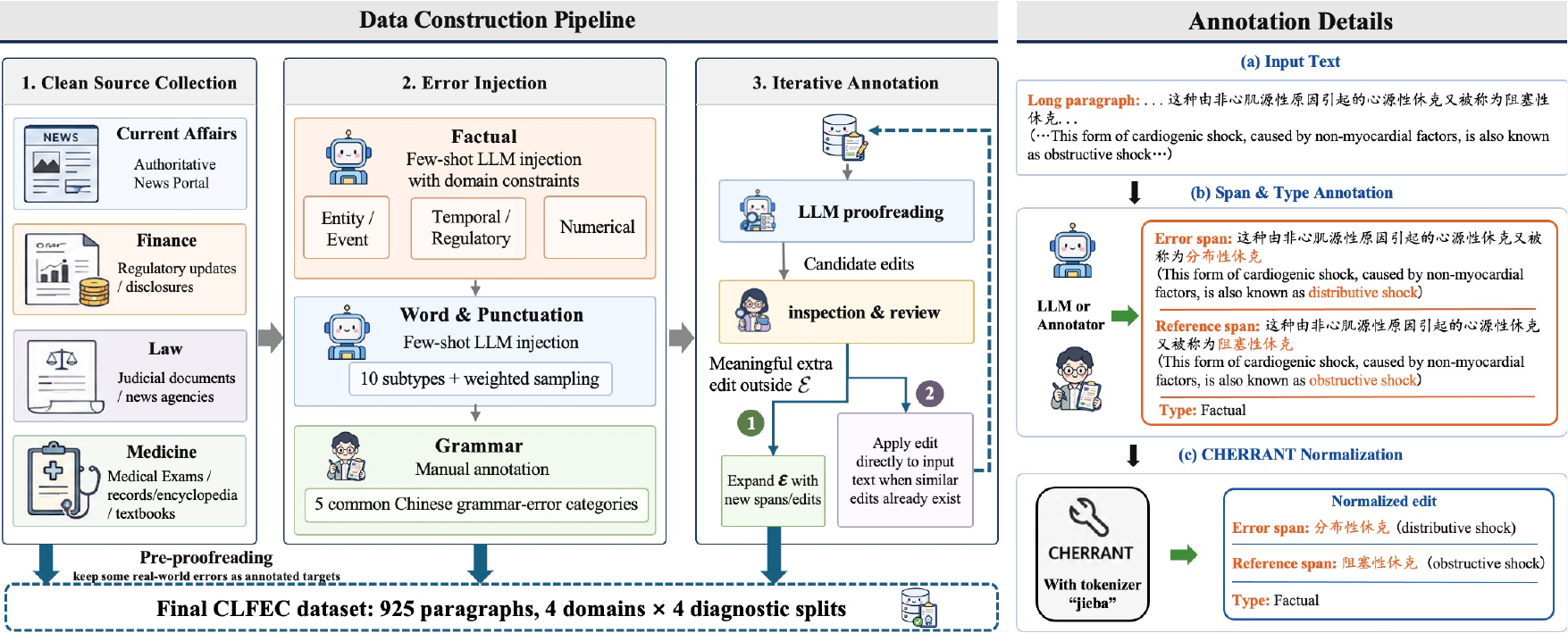}
  \caption{Overview of the CLFEC data construction and annotation process. We collect clean professional texts from four domains, inject factual and linguistic errors under controlled settings, iteratively verify candidate edits, and convert span-level annotations into CHERRANT-style normalized edit pairs for evaluation.}
  \label{fig:data_construction}
  \vspace{-14pt}
\end{figure*}

\section{The CLFEC Task}
\subsection{Task Definition}
In CLFEC, we jointly consider LEC and FEC in a unified setting. Given a long-form professional Chinese paragraph $x=(c_{1},\ldots,c_{n})$ with $c_i \in \mathcal{V}$ (character vocabulary $\mathcal{V}$), the task is to generate a corrected paragraph $y=(d_{1},\ldots,d_{m})$ with $d_j \in \mathcal{V}$. Since edits may involve insertions and deletions, $n$ and $m$ need not be equal.

The ground truth annotation is represented as a set of \emph{edit operations} $\mathcal{E}=\{(s_k, e_k, u_k, v_k)\}_{k=1}^{K}$. For each edit $k$, $s_k$ and $e_k$ denote the start and end indices in the source text, $u_k = x_{[s_k:e_k]}$ represents the original erroneous span, and $v_k$ represents the corrected target span. A task example instance is detailed in Appendix~\ref{app:task-sample}. The model restricts modifications to regions defined by $\mathcal{E}$, ensuring that $\mathrm{Diff}(x,y)$ aligns with the intended edits. The output $y$ should restore readability and factual consistency while preserving the meaning of non-editable regions. When multiple valid repairs exist, the model should prefer minimal necessary changes. For example, in medical narratives where the chief complaint contradicts the diagnosis, the system should resolve the inconsistency by applying the smallest set of edits (e.g., correcting only the entity rather than rewriting the entire description).

\subsection{Task Dataset}
As illustrated in Figure~\ref{fig:data_construction}, our dataset is constructed from high-quality sources via multi-stage error injection, followed by iterative annotation and manual verification.

\paragraph{Data Source} We collect long-form texts from four specialized domains: current affairs, finance, law, and medicine. Each domain is curated and validated by a dedicated domain expert annotator. 
For current affairs, we crawl news articles from major mainstream media portals.\footnote{\url{https://www.people.com.cn/}}
For finance, we collect regulatory updates, penalty announcements, and market commentaries from professional financial media and disclosure platforms.\footnote{\url{https://www.financeun.com/}, \url{https://www.eastmoney.com/}}
For law, we sample judicial documents and related materials from public repositories and authoritative news agencies.\footnote{\url{https://wenshu.court.gov.cn/}, \url{https://www.xinhuanet.com/}}
For medicine, we sample from publicly available resources such as multiple-choice examinations~\cite{wang-etal-2024-cmb}, de-identified electronic medical records~\cite{li2025chinesedischargedrugrecommendation}, medical encyclopedia Q\&A~\cite{li2023huatuo26m}, and medical textbooks. Even with strict control over data sources, the collected texts still contain artifacts such as malformed mathematical symbols, OCR recognition errors, and occasional colloquial expressions. We run a pre-proofreading pass and intentionally keep a subset of these issues as annotated targets, which makes the dataset closer to real-world inputs.

\paragraph{Error Injection} We build the dataset via error injection with LLM, followed by manual verification. Each paragraph contains at most one injected factual error, and we inject linguistic errors at a fixed density of one error per 150 Chinese characters. Factual errors are injected as localized inconsistencies, including entity/event substitution (e.g., incorrect person/organization/event name), temporal or regulatory mismatch (e.g., outdated policy reference), and numerical inconsistency (e.g., wrong count/rate/date). Word and punctuation errors are injected automatically with subtype sampling (Table~\ref{app:sle_weights}). Grammatical errors are annotated manually due to the low controllability of LLM-generated corruptions, covering five common Chinese grammar-error categories: improper word or clause order, improper collocation, missing required components, redundant components, and structurally mixed or incoherent sentences.

For each injected error, we first annotate a span-level correction triple consisting of an \textit{error span}, a \textit{reference span}, and an error type.  We then normalize each span-level annotation into a minimal edit following \textsc{ChERRANT}~\cite{zhang-etal-2022-mucgec} with the \textit{jieba}\footnote{\url{https://github.com/fxsjy/jieba}} tokenizer, yielding the final editable edit set $\mathcal{E}$. The corrupted paragraph is produced by applying $\mathcal{E}$ to the original paragraph.
\label{sec:data_construction}

\paragraph{Iterative annotation}
We refine annotations through iterative LLM proofreading and manual inspection. When model predictions reveal meaningful edits outside the current edit set $\mathcal{E}$, annotators either add the corresponding span-level correction to $\mathcal{E}$ or, for recurrent issues already sufficiently represented, apply the correction to both the input and gold text.

\begin{figure}[t]
  \includegraphics[width=0.9\columnwidth]{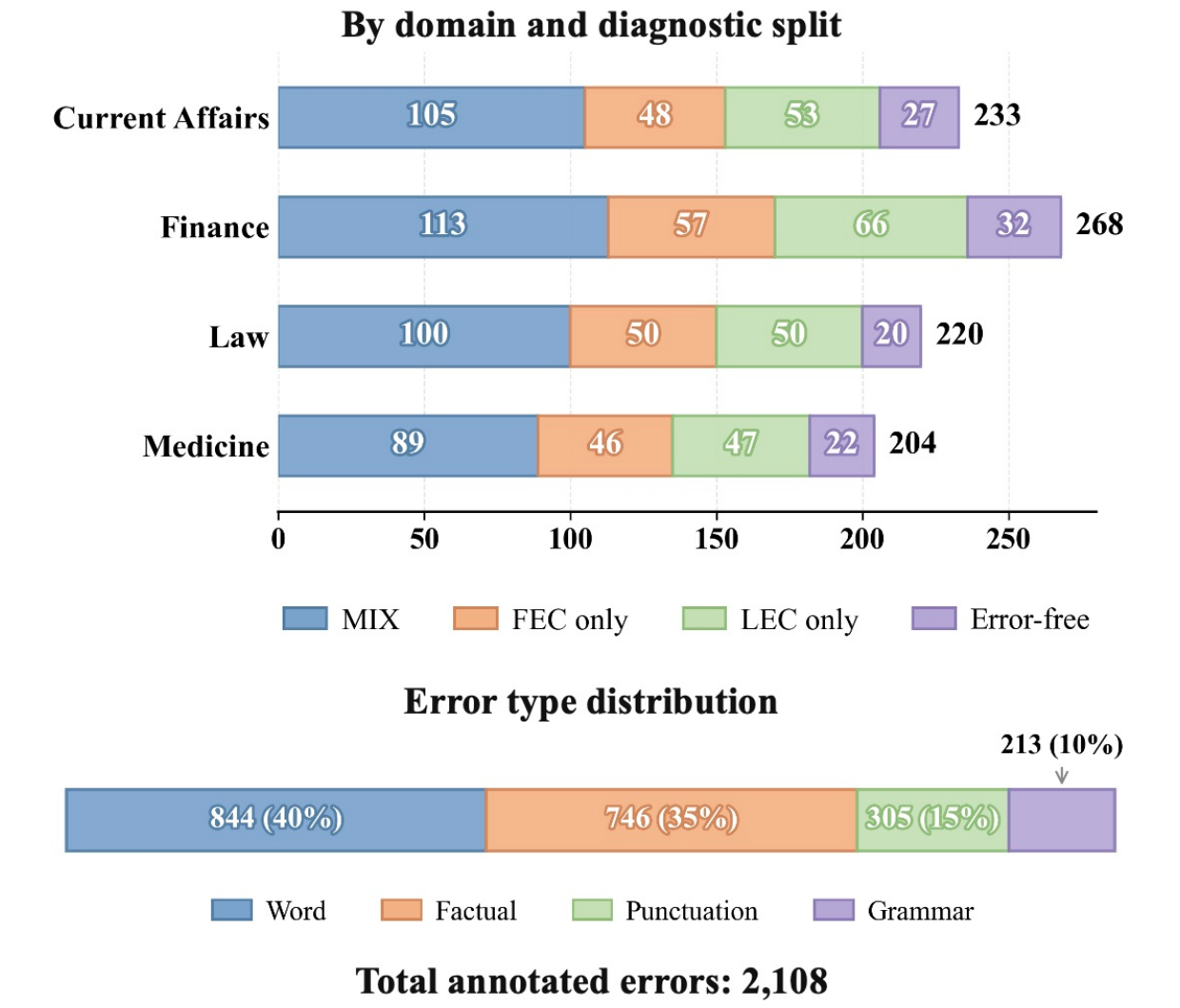}
  \caption{Statistics of the CLFEC dataset.}
  \vspace{-14pt}
  \label{fig:dataset}
\end{figure}

\paragraph{Dataset Statistics}
Figure~\ref{fig:dataset} summarizes the statistics of CLFEC, including domain distribution, diagnostic splits, and error-type distribution. In the diagnostic split design, we allocate a larger proportion to MIX, as mixed linguistic--factual cases are the core scenario of CLFEC. Overall, the dataset contains \textbf{430,706} Chinese characters with an average paragraph length of \textbf{465.63} characters. The overall error density is \textbf{48.94} errors per 10K characters, with an average of \textbf{2.56} errors per erroneous paragraph. 

\subsection{Human Evaluation}
To evaluate whether CLFEC errors resemble naturally occurring writing mistakes, we conduct a human evaluation of error naturalness. Annotators are shown Chinese texts containing errors and are asked to judge whether each error is closer in style to a naturally occurring error or to an artificially/model-constructed synthetic error. To avoid confounding error naturalness with error-type distribution, we adopt a type-balanced sampling strategy. Specifically, we sample 50 instances for each error type from both CLFEC and real-world samples (RWS), yielding 200 CLFEC samples and 200 real-world samples in total. Two annotators (A1 and A2) independently label each sample. Table~\ref{tab:human-naturalness} reports the annotation results.

\begin{table}[t]
\centering
\small
\setlength{\tabcolsep}{4pt}
\begin{tabular}{llccc}
\hline
\textbf{Dataset} & \textbf{Ann.} & \textbf{Natural} & \textbf{Synthetic} & \textbf{Agreement} \\
\hline
\multirow{2}{*}{CLFEC} 
  & A1 & 143 (71.5\%) & 57 (28.5\%) & \multirow{2}{*}{84.0\%} \\
  & A2 & 143 (71.5\%) & 57 (28.5\%) &  \\
\hline
\multirow{2}{*}{RWS} 
  & A1 & 179 (89.5\%) & 21 (10.5\%) & \multirow{2}{*}{80.0\%} \\
  & A2 & 159 (79.5\%) & 41 (20.5\%) &  \\
\hline
\end{tabular}
\caption{Human evaluation of error naturalness. Agreement denotes the percentage of samples on which the two annotators assign the same label.}
\vspace{-14pt}
\label{tab:human-naturalness}
\end{table}

As shown in Table~\ref{tab:human-naturalness}, under type-balanced sampling, both annotators judge the majority of CLFEC errors as natural, with the same naturalness rate of 71.5\% and an inter-annotator agreement of 84.0\%. The real-world Sample set receives higher naturalness rates, 89.5\% and 79.5\% from the two annotators, with 80.0\% agreement. Although real-world samples receive higher naturalness scores, CLFEC still obtains a high naturalness rate and comparable annotator agreement.

\section{Proofreading Pipelines}

\label{sec:method}
CLFEC requires joint correction of linguistic and factual errors at paragraph level. We investigate two primary architectural paradigms: retrieval-augmented pipelines and an agentic workflow, both built on LLMs with external evidence retrieval.

\subsection{Retrieval-Augmented Proofreading}
\label{sec:pipelines}

\paragraph{RAG-based Module}
It consists of three steps: (1) Scanning: The LLM scans the input text and optionally generates up to three search queries for uncertain claims; (2) Retrieval: Relevant evidence is retrieved using a \texttt{search\_tool}; (3) Correction: The model generates the final corrected text conditioned on the input and the evidence. 

Based on this module, we implement two variations: (1) \textbf{Sequential RAG (S-RAG):} To mirror the common sequential workflow in industrial applications, we construct a two-stage process. In Stage 1, the LLM is prompted to perform LEC only to normalize the text without accessing external tools. In Stage 2, the Stage 1 output is passed to the RAG Module, which focuses only on factual error correction. (2) \textbf{Unified RAG (U-RAG):} This variant handles LEC+FEC in a single RAG pass, using a unified prompt to identify and correct both error types within the same context.

\paragraph{External Knowledge Acquisition}
We implement \texttt{search\_tool} to construct structured evidence based on the \textit{Quark}~\footnote{\url{https://vt.quark.cn/blm/qk-ai-business-page-915/index?x_render_type=stream_ssr}} Business Search API. Upon receiving a query, the tool retrieves a set of candidate references. To filter noise and fit the context window, the results are re-ranked via BM25~\cite{10.1561/1500000019} to retain the top-3 snippets, each truncated to the first 512 characters. Each evidence item includes a title, a link, a timestamp, and the snippet content. \label{sec:search_tool}

\subsection{Agentic Proofreading Framework}
\label{sec:agent}
The agentic framework operates under a ReAct paradigm~\cite{yao2023reactsynergizingreasoningacting}. Through meticulous planning and execution, and multiple rounds of feedback verification, it aims to improve the precision and recall of proofreading.

\paragraph{Plan-and-Execute}
To mitigate long-context issues (e.g., lost-in-the-middle~\cite{liu-etal-2024-lost}), we instructed the agents to use a state management tool \texttt{todo\_write} to follow a strict “plan and execute” workflow, explicitly limiting the agents to focus on specific target locations in each round of interaction. Upon receiving the document, the agent first scans the document and writes a localized to-do list covering linguistic checks and factual verification tasks. It then processes tasks sequentially, either editing directly or invoking \texttt{search\_tool} for factual claims. The state-management tool \texttt{todo\_write} tracks task status and helps keep edits localized and traceable.


\paragraph{Verification and Feedback}
To mitigate hallucination, we use a deterministic \texttt{verify\_tool} that enforces a unique anchor span in the source text. Crucially, the tool returns targeted error messages for common failure cases (e.g., missing or no-op anchors, appearing multiple times), which enables the agent to self-correct in subsequent turns. Validated edits are then stored, so the final output contains only executable and unambiguously grounded corrections.

\begin{table*}[t]
  \centering
  \scriptsize
  \setlength{\tabcolsep}{3.5pt}
  \renewcommand{\arraystretch}{1.08}

  \resizebox{\textwidth}{!}{%
  \begin{tabular}{l l ccc ccc ccc ccc ccc}
    \toprule
    \multirow{2}{*}{\textbf{model}} &
    \multirow{2}{*}{\textbf{method}} &
    \multicolumn{3}{c}{\textbf{Detection}} &
    \multicolumn{3}{c}{\textbf{Correction}} &
    \multicolumn{3}{c}{\textbf{Correction (MIX)}} &
    \multicolumn{3}{c}{\textbf{Correction (LEC)}} &
    \multicolumn{3}{c}{\textbf{Correction (FEC)}} \\
    \cmidrule(lr){3-5}\cmidrule(lr){6-8}\cmidrule(lr){9-11}\cmidrule(lr){12-14}\cmidrule(lr){15-17}
    & & \textbf{P} & \textbf{R} & \textbf{F1}
      & \textbf{P} & \textbf{R} & \textbf{F1}
      & \textbf{P} & \textbf{R} & \textbf{F1}
      & \textbf{P} & \textbf{R} & \textbf{F1}
      & \textbf{P} & \textbf{R} & \textbf{F1} \\
    \midrule

    \multicolumn{17}{c}{\textbf{LLM-based systems}} \\
    \midrule

    CEC3-4B
      & prompt
      & 34.06 & 14.90 & 20.73
      & 31.78 & 14.04 & 19.47
      & 39.61 & 13.62 & 20.27
      & 38.95 & 21.58 & 27.77
      & -- & -- & -- \\
    \midrule

    \multirow{3}{*}{Qwen3-4B}
      & prompt
      & 37.74 & 23.15 & 28.70
      & 31.71 & 20.20 & 24.68
      & 43.52 & 18.93 & 26.39
      & 37.74 & 30.43 & 33.69
      & 6.97 & 7.14 & 7.05 \\
      & S-RAG
      & 35.72 & 47.01 & 40.60
      & 31.40 & 43.81 & 36.58
      & 40.07 & 42.92 & 41.45
      & 32.32 & 44.13 & 37.32
      & 19.89 & 47.86 & 28.11 \\
      & U-RAG
      & 43.17 & 39.85 & 41.44
      & 39.16 & 37.54 & 38.33
      & 49.68 & 35.63 & 41.50
      & 39.33 & 39.75 & 39.54
      & 28.07 & 44.02 & 34.28 \\
    \midrule

    \multirow{4}{*}{DS-V3.2}
      & prompt
      & 38.54 & 48.24 & 42.85
      & 34.94 & 45.80 & 39.64
      & 44.66 & 43.77 & 44.21
      & 47.57 & 64.61 & 54.79
      & 44.66 & 43.77 & 44.21 \\
      & S-RAG
      & 32.55 & 68.98 & 44.23
      & 30.06 & 67.25 & 41.55
      & 39.84 & 65.21 & 49.46
      & 32.00 & \underline{72.79} & 44.46
      & 17.61 & 67.53 & 27.93 \\
      & U-RAG
      & 45.49 & 67.50 & 54.35
      & 43.13 & 66.32 & 52.27
      & 50.94 & 63.49 & 56.53
      & 47.55 & 71.37 & 57.07
      & 34.21 & 71.91 & 46.36 \\
      & Agent
      & 55.18 & \textbf{71.30} & 62.21
      & 53.01 & \textbf{70.47} & 60.51
      & 64.36 & \textbf{69.31} & 66.74
      & 58.43 & 72.36 & 64.66
      & 37.07 & 73.19 & 49.21 \\
    \midrule

    \multirow{4}{*}{Qwen3-235B}
      & prompt
      & 38.49 & 38.00 & 38.24
      & 33.16 & 34.55 & 33.84
      & 42.34 & 32.80 & 36.96
      & 44.51 & 50.78 & 47.44
      & 8.91 & 13.19 & 10.63 \\
      & S-RAG
      & 23.35 & 61.95 & 33.92
      & 20.98 & 59.39 & 31.00
      & 28.06 & 57.23 & 37.66
      & 23.18 & 62.26 & 33.78
      & 13.21 & 65.91 & 22.00 \\
      & U-RAG
      & 26.37 & 62.48 & 37.08
      & 24.28 & 60.53 & 34.66
      & 33.48 & 58.73 & 42.65
      & 23.25 & 60.64 & 33.61
      & 17.86 & 70.48 & 28.50 \\
      & Agent
      & 25.08 & 55.46 & 34.53
      & 22.63 & 52.91 & 31.70
      & 32.25 & 51.77 & 39.74
      & 26.05 & 54.26 & 35.20
      & 11.56 & 56.50 & 19.19 \\
    \midrule

    \multirow{4}{*}{GLM-4.7}
      & prompt
      & 50.42 & 54.74 & 52.49
      & 46.09 & 52.51 & 49.09
      & 55.89 & 50.34 & 52.97
      & 55.52 & 65.60 & 60.14
      & 24.44 & 38.10 & 29.78 \\
      & S-RAG
      & 52.30 & \underline{70.16} & 59.93
      & 48.87 & 68.72 & 57.12
      & 57.20 & 66.16 & 61.35
      & 50.29 & \textbf{74.14} & 59.93
      & 38.69 & 72.15 & 50.37 \\
      & U-RAG
      & 66.03 & 69.97 & \textbf{67.94}
      & 62.89 & \underline{68.94} & \textbf{65.78}
      & 72.08 & \underline{66.39} & \textbf{69.12}
      & 64.24 & 71.64 & \underline{67.74}
      & \underline{52.56} & \textbf{77.73} & \textbf{62.71} \\
      & Agent
      & \textbf{67.84} & 60.34 & 63.87
      & \underline{64.80} & 59.24 & 61.90
      & \textbf{74.10} & 56.71 & 64.25
      & \textbf{74.19} & 62.85 & \textbf{68.05}
      & 43.41 & 66.67 & 52.58 \\
    \midrule

    \multirow{4}{*}{Kimi-K2.5}
      & prompt
      & 48.61 & 53.80 & 51.07
      & 44.92 & 51.83 & 48.13
      & 56.50 & 50.04 & 53.07
      & 54.90 & 67.38 & 60.50
      & 18.18 & 30.77 & 22.86 \\
      & S-RAG
      & 49.52 & 67.98 & 57.30
      & 46.86 & 66.77 & 55.07
      & 56.96 & 64.58 & 60.53
      & 48.00 & 71.04 & 57.29
      & 33.88 & 70.39 & 45.75 \\
      & U-RAG
      & \underline{67.31} & 66.60 & \underline{66.95}
      & \textbf{64.86} & 65.78 & \underline{65.31}
      & \underline{72.66} & 64.95 & \underline{68.59}
      & 64.24 & 66.67 & 65.43
      & \textbf{55.56} & 68.46 & \underline{61.34} \\
      & Agent
      & 64.72 & 67.88 & 66.27
      & 62.19 & 67.01 & 64.51
      & 71.82 & 65.24 & 67.63
      & \underline{68.06} & 67.85 & 64.88
      & 49.58 & \underline{74.58} & 59.57 \\
    \midrule

    \multicolumn{17}{c}{\textbf{Commercial proofreading products}} \\
    \midrule

    P1
      & \multicolumn{1}{c}{--}
      & 25.18 & 42.41 & 31.60
      & 22.64 & 39.84 & 28.87
      & 28.42 & 38.25 & 32.61
      & 32.71 & 60.34 & 42.42
      & 3.95 & 8.79 & 5.45 \\
    P2
      & \multicolumn{1}{c}{--}
      & 28.96 & 24.91 & 26.78
      & 25.21 & 22.40 & 23.72
      & 31.51 & 22.99 & 26.59
      & 28.29 & 30.74 & 29.46
      & 4.25 & 3.64 & 3.92 \\

    P3
      & \multicolumn{1}{c}{--}
      & 24.18 & 27.04 & 25.53
      & 20.49 & 23.90 & 22.06
      & 24.88 & 23.22 & 24.02
      & 25.16 & 33.12 & 28.60
      & 7.28 & 9.66 & 8.30 \\

    \bottomrule

  \end{tabular}%
  }

  \caption{\label{tab:all-results}
    Overall results on \textbf{CLFEC}. We report Precision (P), Recall (R), and F1 for detection and correction, and further break down correction performance on three subsets: \textbf{MIX}, \textbf{LEC}, and \textbf{FEC}. Best and second-best numbers are highlighted in bold and underline, respectively.
  }
  \vspace{-14pt}
\end{table*}

\section{Experiments}
\subsection{Experimental Setup}
We employ \textbf{ChineseErrorCorrector3-4B (CEC3-4B)}~\cite{tian2025chineseerrorcorrector34bstateoftheartchinesespelling} for the LEC task, using its backbone model \textbf{Qwen3-4B}~\cite{yang2025qwen3technicalreport} for comparison. CEC3-4B was trained on large-scale data tailored for CSC and GEC and achieves state-of-the-art performance on benchmark tests. For all LLM-based experiments, we employ four representative large language models as backbones: \textbf{DeepSeek-V3.2 (DS-V3.2)}~\cite{deepseekai2025deepseekv32pushingfrontieropen}, \textbf{Qwen3-235B}~\cite{yang2025qwen3technicalreport}, \textbf{Kimi-K2.5}~\cite{kimiteam2026kimik25visualagentic}, and \textbf{GLM-4.7}~\cite{5team2025glm45agenticreasoningcoding}. To ensure fair comparisons, we maintained consistent prompt templates and the same output format across all experimental setups (Appendix~\ref{app:prompts_by_method}). The temperature is set to the lowest level of 0.01 to ensure stability. We also include the commercial proofreading products introduced in Section~\ref{sec:motivation} as real-world system baselines.

\subsection{Metrics}
Following the \textsc{ChERRANT}~\cite{zhang-etal-2022-mucgec} paradigm, system performance is evaluated by measuring the alignment between predicted edits and the ground-truth edits. To account for the equal importance of precision and recall in our setting, we adopt word-level F1-scores instead of the conventional F$_{0.5}$ metric. We report both the detection and correction results and perform strict matching, which requires that the error location and candidate span be the same as the annotation.

\subsection{Key Challenges}

\begin{figure}[t]
  \includegraphics[width=0.9\columnwidth]{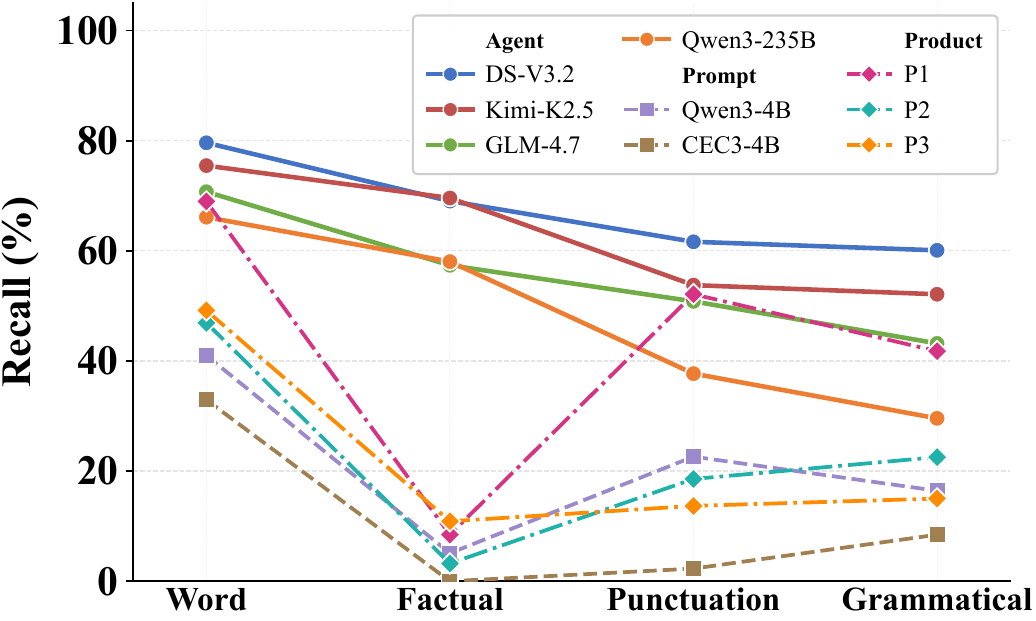}
  \caption{Detection recall by error type on CLFEC.}
  \vspace{-14pt}
  \label{fig:error_recall}
\end{figure}

\paragraph{Specialized model underperforms general LLMs on LEC task.}
We evaluated CEC3-4B using both single-sentence and whole-paragraph calling strategies, reporting the best results. As shown in Table~\ref{tab:all-results}, the CEC3-4B underperforms the Qwen3-4B on the LEC task. Qwen3-4B achieves a Recall of 30.43\% and an F1 score of 33.69\%, outperforming CEC3-4B by approximately \textbf{9\%} and \textbf{6\%}, respectively. Their performance across three types of linguistic errors is further examined in Figure~\ref{fig:error_recall}. CEC3-4B shows almost no ability to correct punctuation errors, and its word and grammatical corrections also lag behind the general model, despite being the focus of specialized training. These results suggest that current dedicated error correction models do not transfer reliably to new domains, whereas foundation models are more robust.

\paragraph{FEC requires retrieval and evidence grounding.}
To validate the effectiveness of our constructed FEC dataset and evaluate the model's ability to correct facts, we compared prompt-only baselines with three external evidence injection methods. On the four large-parameter models, pure prompts only achieved an average \textbf{Precision of 24.05\%} and \textbf{Recall of 31.46\%}, resulting in a sub-optimal F1 score of 26.87\%. In contrast, the U-RAG method improved the average F1 score to 49.73\%, with GLM-4.7 and Kimi-K2.5 seeing gains of over 30 percentage points. This performance gap confirms that parameter knowledge alone cannot reliably solve the fact correction problem, as the errors injected are often specific or timely, necessitating explicit evidence for accurate correction.

\paragraph{Mixed-error inputs present the greatest challenge.}
Table~\ref{tab:all-results} shows that the MIX split yields the lowest Recall across all large-parameter models, compared with splits that contain only one error type. For instance, even the best-performing GLM-4.7 (U-RAG) drops from \textbf{77.73\%} (FEC) and \textbf{71.64\%} (LEC) to \textbf{66.39\%} in the MIX setting. Our qualitative analysis reveals a \textbf{``masking effect''}: when a linguistic error is layered on top of a factual error, models often fix the surface form but miss the underlying inconsistency. A typical medical example involves the input \zh{“...诊断为尿到炎”} (``...diagnosed as urethritis'' with a typo). Models often correct it to \zh{“...诊断为尿道炎”} (``...diagnosed as urethritis'' with correct spelling), satisfying the language model's fluency constraint, yet failing to recognize that the symptoms actually point to \zh{“...诊断为阴道炎”} (``...diagnosed as vaginitis'').

\paragraph{Punctuation and grammar are more difficult.} 
Figure~\ref{fig:error_recall} reports detection recall by error type, covering agent-based models, prompt-only baselines, and commercial products. Among the agent-based models, grammatical and punctuation errors are recalled less reliably than word and factual errors. The product P1 can still capture part of these cases through rule-based checks, leading to competitive recall on punctuation and grammatical errors. In addition, we find that these two error types contain a large number of \textbf{``weakly normative''} issues. Native speakers often overlook these errors without careful checking against explicit rules. For example, parallel quotation marks and book titles cannot be connected by a comma (\zh{、}), and the phrase \zh{“开始启用”} (``begin start using'') is semantically redundant. This gray area can lead to both undercorrection and overcorrection (unnecessary normalization), which makes a conservative and auditable editing strategy important. Meanwhile, products such as P1 can still achieve competitive recall on these errors through rule-based methods.

\subsection{Results and Analysis}
\paragraph{Over-correction of LLMs}
\label{sec:overcorrection}
We observed a clear pattern: precision increases with sample error density. Across the MIX, LEC, and FEC splits, error density decreases in that order, and Table~\ref{tab:all-results} shows a matching drop in precision for all methods. This pattern appears as higher precision on error-dense inputs, but the highest average false-positive rate per sample on the Error-free split. When no obvious errors are present, LLMs often revise style or preference-based wording even under explicit constraints. These changes are commonly justified with vague rationales, such as \zh{“更正式”} (``more formal'') or \zh{“更标准”} (``more standard''). This indicates that LLMs lack a stable "standard reference" in the "should it be changed?" step, and will actively seek out points for modification in order to complete the "error correction" task, ultimately leading to over-correction.

\paragraph{Benefit of Unified Context}
Comparing the S-RAG with Unified approaches, we find that the two-stage strategy severely degrades precision. There are two factors:
(1) \textbf{Accumulative Over-correction:} Conducting correction twice amplifies the overcorrection effect of the model. As discussed above, the model makes more style modifications on a "cleaner text", thus deviating from the error correction principle.
(2) \textbf{Contextual Misalignment:} Initial linguistic edits often alter sentence structures and token indices. Even if the subsequent factual module correctly retrieves evidence, it frequently fails to map the correction back to the modified text, leading to alignment failures. 
Beyond precision, unified correction also improves efficiency: U-RAG requires one fewer LLM call than S-RAG.

Contrary to the hypothesis that factual retrieval could aid linguistic correction, our results show that U-RAG does not improve LEC recall in the unified setting. One likely reason is that the retrieved evidence is optimized for factual verification and often does not align with the contextual cues needed for LEC. In practice, linguistic disambiguation is usually resolved by the paragraph context itself.

\paragraph{Agent Performance}
As shown in Table~\ref{tab:all-results}, the effectiveness of the agentic framework depends on the backbone model's agentic capability. With DeepSeek-V3.2, the Agent achieves higher F1 for both detection and correction. Specifically, it improves FEC Recall by +1.28 points over U-RAG while maintaining higher precision (+2.86 points), suggesting that the model can reliably plan, execute, and validate. In contrast, for Qwen3-235B and GLM-4.7, the agentic framework slightly underperforms its RAG counterpart, possibly due to both the agent workflow and model-specific behaviors that make multi-turn interactions more prone to instruction drift and formatting errors.

Analysis further suggests that the Agent's main advantage over RAG-based methods is its ability to autonomously resolve \textbf{conflicting evidence}.  For instance, in a case involving government statistics, the Agent explicitly reasoned: \emph{``Search results show data inconsistency. Results 1 and 3 support the original text `National level 18, Autonomous region level 136, City level 275', but Result 2 shows `Autonomous region level 123'. I need to further verify which data is more accurate.''} By triggering a targeted second search, the Agent successfully identified the outdated source and preserved the correct original text.

\paragraph{Commercial proofreading products.}
The commercial proofreading products exhibit limitations consistent with the analysis in Section~\ref{sec:motivation}. They have generally low precision, and their FEC performance is uniformly weak, with both precision and recall below 10\%. As shown in Figure~\ref{fig:error_recall}, the best-performing P1 obtains reasonable recall across word, grammar, and punctuation, but it remains insufficient for FEC. These results indicate that CLFEC is challenging for existing proofreading products.

\section{Related Work}


Existing Chinese correction benchmarks mainly evaluate linguistic well-formedness. MuCGEC addresses multi-source and multi-reference Chinese GEC~\citep{zhang-etal-2022-mucgec}, NaSGEC extends CGEC to native-speaker texts from multiple domains~\citep{zhang-etal-2023-nasgec}, and CCTC further emphasizes cross-sentence context~\citep{wang-etal-2022-cctc}. These benchmarks provide useful testbeds for spelling, grammar, and word-usage correction, but they do not require systems to verify factual claims or ground corrections in external evidence.

A separate line of work studies correction in knowledge-intensive settings. MEDEC focuses on clinically relevant errors in medical notes~\citep{ben-abacha-etal-2025-medec}, while Chinese government text correction incorporates external knowledge bases to handle terminology, policy names, and institutional expressions~\citep{kartc-2025}. These studies are closer to professional proofreading, but they are typically domain-specific and do not provide a unified setting where linguistic and factual errors co-occur in paragraph-level Chinese writing. CLFEC differs from both lines by evaluating linguistic correction, factual correction, mixed-error correction, and over-correction within one diagnostic benchmark.

\section*{Conclusion}
We present CLFEC, a new paragraph-level task for unified linguistic and factual error correction in Chinese professional writing. Motivated by real corpus analysis, we construct and release a mixed, multi-domain dataset through controlled error injection on verified professional texts, with diagnostic LEC, FEC, MIX, and Error-free splits. We further evaluate proofreading pipelines spanning prompting, RAG, and agentic workflows. Results show that evidence grounding is crucial for factual repair, mixed-error inputs remain challenging, and unified correction outperforms decoupled pipelines. Overall, CLFEC provides a new benchmark for Chinese text correction research and practical guidance for proofreading systems.

\section*{Limitations}
First, the evaluation mainly relies on automated metrics with strict span matching against a single gold reference. Although the dataset is annotated under rigorous quality control, paragraph-level proofreading, especially for factual consistency and stylistic polishing, is inherently open-ended. A single reference cannot cover all valid corrections, so valid but non-matching edits may be penalized, underestimating model performance. Future work will add multi-reference annotations and incorporate human evaluation to better reflect real-world acceptability.

Second, high-performing methods on CLFEC evaluation, particularly the agentic framework, currently depend on large-scale foundation models, which incur significant inference latency and computational costs. In our experiments, smaller models (e.g., 4B parameters) still struggle with the complex reasoning and tool-use. Exploring low-resource methods—such as knowledge distillation or lightweight agentic frameworks—to enhance the capability of small models for real-time industrial applications remains a critical direction for future research.

\bibliography{custom}

\clearpage
\appendix
\section{Additional Results}
\subsection{Detailed Error Recall Analysis}
\label{app:detailed_recall}
In this section, we provide the comprehensive breakdown of recall rates across four specific error types: Word, Factual, Punctuation and Grammatical. Table~\ref{tab:error-recall-all} supplements the aggregated results presented in the main text (Figure~\ref{fig:error_recall}), detailing the performance of various backbone models under different methods (Prompt-only, S-RAG, U-RAG, and Agent).

\begin{table}[t]
  \centering
  \scriptsize
  \setlength{\tabcolsep}{4pt}
  \renewcommand{\arraystretch}{1.08}
  \begin{tabular}{l l cccc}
    \toprule
    \textbf{Model} & \textbf{Method} & \textbf{Word} & \textbf{Factual} & \textbf{Punctuation} & \textbf{Grammatical} \\
    \midrule

\multicolumn{6}{c}{\textbf{LLM-based systems}} \\
  \midrule
  CEC3-4B & prompt & 32.94 & 0.00 & 2.30 & 8.45 \\
  \midrule

  \multirow{3}{*}{Qwen3-4B}
    & prompt & 41.00 & 5.09 & 22.62 & 16.43 \\
    & S-RAG  & 59.00 & 44.77 & 32.79 & 27.70 \\
    & U-RAG  & 47.51 & 41.96 & 24.92 & 23.47 \\
  \midrule

  \multirow{4}{*}{DS-V3.2}
    & prompt & 70.02 & 19.44 & 59.02 & 47.42 \\
    & S-RAG  & 74.29 & 67.69 & \textbf{62.62} & \textbf{61.50} \\
    & U-RAG  & 74.88 & 66.22 & 57.38 & \underline{57.28} \\
    & Agent  & \textbf{79.62} & 69.03 & \underline{61.64} & 60.09 \\
  \midrule

  \multirow{4}{*}{Qwen3-235B}
    & prompt & 62.91 & 10.99 & 41.31 & 29.11 \\
    & S-RAG  & 69.08 & 64.61 & 48.20 & 44.13 \\
    & U-RAG  & 68.48 & 66.22 & 47.87 & 46.48 \\
    & Agent  & 66.11 & 58.04 & 37.70 & 29.58 \\
  \midrule

  \multirow{4}{*}{GLM-4.7}
    & prompt & 73.10 & 34.85 & 57.38 & 47.89 \\
    & S-RAG  & \underline{76.90} & \underline{70.91} & \underline{61.64} & 53.05 \\
    & U-RAG  & 76.07 & \textbf{71.98} & 60.00 & 53.05 \\
    & Agent  & 70.73 & 57.37 & 50.82 & 43.19 \\
  \midrule

  \multirow{4}{*}{Kimi-K2.5}
    & prompt & 75.24 & 28.82 & 60.00 & 47.42 \\
    & S-RAG  & 74.41 & 68.50 & 58.36 & 54.46 \\
    & U-RAG  & 74.53 & 65.82 & 55.41 & 53.99 \\
    & Agent  & 75.47 & 69.59 & 53.77 & 52.11 \\
  \midrule

  \multicolumn{6}{c}{\textbf{Commercial proofreading products}} \\
  \midrule

  P1
    & \multicolumn{1}{c}{--}
    & 68.96 & 8.47 & 52.12 & 41.78 \\

  P2
    & \multicolumn{1}{c}{--}
    & 46.92 & 3.23 & 18.57 & 22.54 \\

  P3
    & \multicolumn{1}{c}{--}
    & 49.17 & 10.89 & 13.68 & 15.02 \\

  \bottomrule
    
  \end{tabular}
  \caption{\label{tab:error-recall-all}
  Recall (\%) for each error type across models and strategies. Best in each column is bold; second best is underlined.
  }
\end{table}

\subsection{Quantitative Analysis of Over-Correction}
\label{app:overcorrection_analysis}
To quantify the model's tendency for over-correction, we introduce the \textbf{Edit Rate} metric, defined as the number of edits performed per 100 characters. Table~\ref{tab:edit-rate} compares the edit rates of the Gold Standard against various baselines across four data splits: MIX, LEC, FEC, and Error-free. The Gold Standard data (Row 1) perfectly reflects the intrinsic error density of the dataset: the edit rate strictly follows the order: MIX(0.72)>LEC(0.47)>FEC(0.33)>Error-free(0.00). This serves as the ground truth reference for ideal model behavior.

The \textbf{S-RAG} method exhibits a massive inflation in edit rates across all backbone models. For instance, with DeepSeek-V3.2, the S-RAG method yields an edit rate of \textbf{1.154} on error-free text—meaning it modifies more than 1 character for every 100 characters even when the text is perfect. 

The Agent framework significantly suppresses the tendency for over-correction compared to both S-RAG and U-RAG baselines. In error-free scenarios, the Agent reduces the edit rate by approximately \textbf{64\%} (from 1.094 to 0.391 with DeepSeek-V3.2), approaching the gold standard. We attribute this robustness to a structural advantage: Unlike end-to-end generation models that mix detection and rewriting in a single pass, our \emph{Plan-and-Execute} workflow forces the agent to explicitly list tasks before execution. This global scanning phase filters out subjective stylistic preferences. 

\begin{table}[t]
  \centering
  \scriptsize
  \setlength{\tabcolsep}{4pt}
  \renewcommand{\arraystretch}{1.08}
  \begin{tabular}{l l cccc}
    \toprule
    \textbf{Model} & \textbf{Method} & \textbf{MIX} & \textbf{LEC} & \textbf{FEC} & \textbf{Error-free} \\
    \midrule

    Gold & gold
      & 0.72 & 0.47 & 0.33 & 0.00 \\
    \midrule

    CEC3-4B & prompt
      & 0.48 & 0.52 & 0.43 & 0.34 \\
    \midrule

    \multirow{3}{*}{Qwen3-4B}
      & prompt & 0.24 & 0.26 & 0.13 & 0.23 \\
      & S-RAG  & 0.72 & 0.61 & 0.77 & 0.32 \\
      & U-RAG  & 0.50 & 0.46 & 0.51 & 0.25 \\
    \midrule

    \multirow{4}{*}{DS-V3.2}
      & prompt & 0.68 & 0.62 & 0.61 & 0.46 \\
      & S-RAG  & 1.29 & 1.53 & 1.54 & 1.15 \\
      & U-RAG  & 0.94 & 1.09 & 0.93 & 1.09 \\
      & Agent  & 0.78 & 0.58 & 0.65 & 0.39 \\
    \midrule

    \multirow{4}{*}{Qwen3-235B}
      & prompt & 0.54 & 0.50 & 0.48 & 0.34 \\
      & S-RAG  & 1.38 & 1.19 & 1.49 & 0.95 \\
      & U-RAG  & 1.20 & 1.18 & 1.22 & 0.96 \\
      & Agent  & 1.14 & 0.98 & 1.56 & 0.87 \\
    \midrule

    \multirow{4}{*}{GLM-4.7}
      & prompt & 0.64 & 0.54 & 0.51 & 0.33 \\
      & S-RAG  & 0.87 & 0.89 & 0.84 & 0.52 \\
      & U-RAG  & 0.72 & 0.64 & 0.68 & 0.43 \\
      & Agent  & 0.55 & 0.40 & 0.50 & 0.15 \\
    \midrule

    \multirow{4}{*}{Kimi-K2.5}
      & prompt & 0.63 & 0.57 & 0.56 & 0.33 \\
      & S-RAG  & 0.94 & 0.93 & 0.96 & 0.68 \\
      & U-RAG  & 0.68 & 0.69 & 0.50 & 0.30 \\
      & Agent  & 0.65 & 0.47 & 0.50 & 0.27 \\
    \bottomrule
  \end{tabular}
  \caption{\label{tab:edit-rate}
   The number of edits performed per 100 characters on MIX/LEC/FEC and an Error-free subset. Higher values on Error-free subset indicate stronger over-correction.
  }
\end{table}

\section{Task and Dataset Details}
\subsection{Task Sample Instance}
\label{app:task-sample}

This sample is representative of the Law domain. The instance features three distinct error types: (1) a \textbf{factual error}, where the fine amount ``\zh{三十万元}'' (300,000 RMB) must be corrected to ``\zh{五十万元}'' (500,000 RMB) to align with actual legislative updates (Article 61); (2) a \textbf{word error}, involving the misuse of the structural particle ``\zh{地}'' which should be ``\zh{的}'' to properly modify the subsequent noun phrase; and (3) a \textbf{punctuation error}, where a missing period is inserted before the closing quotation mark to comply with Chinese orthographic standards.\par

\vspace{0.6em}
\setlength{\fboxsep}{6pt}
\setlength{\fboxrule}{0.4pt}
\noindent\fbox{%
\noindent\begin{minipage}{\dimexpr\columnwidth-2\fboxsep-2\fboxrule\relax}
\small
\setlength{\parskip}{0.35em}
\setlength{\parindent}{0pt}

\field{ID:}{\texttt{new\_law.jsonl\#line\_10}}
\field{Type:}{\texttt{MIX}}

\field{Input:}{\zh{2025年10月28日，第十四届全国人民代表大会常务委员会第十八次会议通过《全国人民代表大会常务委员会关于修改〈中华人民共和国网络安全法〉的决定》。根据该决定，将第五十九条改为第六十一条，修改为：“网络运营者不履行本法第二十三条、第二十七条规定的网络安全保护义务的，由有关主管部门责令改正，给予警告，可以处一万元以上五万元以下罚款；拒不改正或者导致危害网络安全等后果的，处五万元以上\err{三十万元}以下罚款，对直接负责的主管人员和其他直接责任人员处一万元以上十万元以下罚款。关键信息基础设施的运营者不履行本法第三十五条、第三十六条、第三十八条、第四十条规定\err{地}网络安全保护义务的，由有关主管部门责令改正，给予警告，可以处五万元以上十万元以下罚款；拒不改正或者导致危害网络安全等后果的，处十万元以上一百万元以下罚款，对直接负责的主管人员和其他直接责任人员处一万元以上十万元以下罚款。有前两款行为，造成大量数据泄露、关键信息基础设施丧失局部功能等严重危害网络安全后果的，由有关主管部门处五十万元以上二百万元以下罚款，对直接负责的主管人员和其他直接责任人员处五万元以上二十万元以下罚款；造成关键信息基础设施丧失主要功能等特别严重危害网络安全后果的，处二百万元以上一千万元以下罚款，对直接负责的主管人员和其他直接责任人员处二十万元以上一百万元以下罚款\err{”}}}

\field{Corrected:}{\zh{...拒不改正或者导致危害网络安全等后果的，处五万元以上\cor{五十万元}以下罚款...关键信息基础设施的运营者不履行本法第三十五条、第三十六条、第三十八条、第四十条规定\cor{的}网络安全保护义务的...对直接负责的主管人员和其他直接责任人员处二十万元以上一百万元以下罚款\cor{。”}}}

\field{Corrections (\texttt{cors}):}{}
\begin{tightitemize}
  \item \texttt{[191,195)} \quad \texttt{Fact\_Error} \quad
        \zh{\err{三十万元}} $\rightarrow$ \zh{\cor{五十万元}}
  \item \texttt{[274,275)} \quad \texttt{Word\_Error} \quad
        \zh{\err{地}} $\rightarrow$ \zh{\cor{的}}
  \item \texttt{[566,567)} \quad \texttt{Punc\_Error} \quad
        \zh{\err{”}} $\rightarrow$ \zh{\cor{。”}}
\end{tightitemize}

\end{minipage}
}

\subsection{Error Injection Procedure}
\label{app:prompt_templates}
\begin{table}[h]
\centering
\small
\setlength{\tabcolsep}{5pt}
\renewcommand{\arraystretch}{1.0}
\begin{tabular}{l c}
\toprule
Subtype & Weight \\
\midrule
Homophonic confusable (\zh{同音字词}) & 0.15 \\
Shape-similar confusable (\zh{形近字词}) & 0.15 \\
Near-phonetic confusable (\zh{近音字词}) & 0.10 \\
Redundancy (\zh{字词冗余}) & 0.15 \\
Omission (\zh{字词缺失}) & 0.15 \\
Permutation (\zh{字词乱序}) & 0.05 \\
Punctuation misuse (\zh{标点误用}) & 0.10 \\
Punctuation redundancy (\zh{标点冗余}) & 0.05 \\
Punctuation omission (\zh{标点缺失}) & 0.05 \\
Context mismatch (\zh{上下文不对应}) & 0.05 \\
\bottomrule
\end{tabular}
\caption{Word/punctuation error subtypes and sampling weights used in our injection prompts.}
\label{app:sle_weights}
\end{table}

We inject linguistic and factual errors by prompting an LLM to produce a span-level \texttt{error\_map} in JSON format, where each item specifies an \texttt{original\_text} anchor (a contiguous substring in the source), a corrupted \texttt{error\_text}, an \texttt{error\_type}, and a short \texttt{position\_context} to facilitate exact localization. For word and punctuation errors, we first sample a subtype according to the probabilities in Table~\ref{app:sle_weights}, then explicitly pass the sampled subtype to the LLM in the prompt so that the model is instructed to inject that specific kind of error in the current generation round.







\begin{promptfigure}[!t]
  {Word \& Punctuation Error Injection Prompt}
  {Word \& Punctuation Error Injection Prompt}
  {fig:prompt-injection}
\begin{CJK*}{UTF8}{gkai}
\PromptRawStart
你是一个专业的文本错误生成器。你的任务是在给定的原始文本中识别可以插入错误的位置，并生成一个error_map来描述这些错误。

## 要求：
1. 你需要在原文中找到适合插入错误的位置
2. 本次需要生成 {num_errors} 处错误
3. 本次需要生成的错误类型为：{error_types}
4. 你必须返回一个JSON格式的error_map，格式如下：

```json
{{
    "error_map": [
        {{
            "original_text": "原文中的片段（需要能在原文中精确定位）",
            "error_text": "替换后的错误片段",
            "error_type": "错误类型",
            "position_context": "该片段前后的上下文（用于精确定位）"
        }}
    ]
}}
```

## 各错误类型的生成规则：
- **同音字词**：将原文中的字替换为同音字，如"的"→"地"、"在"→"再"、"已经"→"以经"
- **形近字词**：将原文中的字替换为形近字，如"己"→"已"、"未"→"末"、"土"→"士"
- **近音字词**：将原文中的字替换为读音相近的字，如"账"→"帐"、"做"→"作"、"权"→"劝"
- **字词冗余**：在原文中添加多余的字词，如"公司的的业务"、"进行了了审核"
- **字词缺失**：删除原文中的某些字词，如"股份有公司"（缺少"限"）、"中国证监会"→"中国证会"
- **字词乱序**：打乱原文中某些字词的顺序，如"股票上市"→"上市股票"、"审议通过"→"通过审议"
- **标点误用**：替换标点符号，如逗号→句号、顿号→逗号、括号混用
- **标点冗余**：添加多余的标点，如"公司，，决定"、"完成。。"
- **标点缺失**：删除必要的标点，如去掉句子间的逗号或句号
- **上下文不对应**: 上下文的内容不一致，例如上文提到的是A公司，错误文本中提到的是B公司

## 重要注意事项：
1. original_text必须是原文中真实存在的连续片段
2. position_context应包含original_text前后各10-20个字符，用于精确定位
3. 错误应该自然，不要过于明显
4. 确保生成的错误数量为 {num_errors} 个
5. 只返回JSON格式的结果，不要有其他说明文字

请分析以下文本并生成error_map：
\PromptRawEnd
\end{CJK*}
\end{promptfigure}

\begin{promptfigure}[p]
  {Word \& Punctuation Error Injection Prompt (English Translation)}
  {Word \& Punctuation Error Injection Prompt (English Translation)}
  {fig:prompt-injection-en}
\PromptRawStart
You are a professional text error generator. Your task is to identify positions in a given source text where errors can be inserted and generate an error_map describing those errors.

## Requirements:
1. You need to find suitable positions in the original text for inserting errors.
2. You need to generate {num_errors} errors in this round.
3. The error types to generate in this round are: {error_types}
4. You must return an error_map in JSON format as follows:

```json
{{
    "error_map": [
        {{
            "original_text": "A span from the original text that can be located precisely",
            "error_text": "The replaced erroneous span",
            "error_type": "Error type",
            "position_context": "Context before and after the span for precise localization"
        }}
    ]
}}
```

## Generation rules for each error type:
- **Homophonic character / word substitution**: replace a character with a homophone, such as replacing one character with another that has the same pronunciation.
- **Visually similar character / word substitution**: replace a character with a visually similar one.
- **Near-pronunciation character / word substitution**: replace a character with one that has a similar pronunciation.
- **Redundant character / word**: add unnecessary characters or words to the original text.
- **Missing character / word**: delete certain characters or words from the original text.
- **Character / word order confusion**: disturb the order of certain words or phrases in the original text.
- **Punctuation misuse**: replace punctuation marks, such as changing a comma into a period or mixing bracket types.
- **Redundant punctuation**: add unnecessary punctuation marks.
- **Missing punctuation**: delete necessary punctuation marks.
- **Context mismatch**: make the content inconsistent with the surrounding context, for example, the previous context mentions Company A while the erroneous text mentions Company B.

## Important notes:
1. original_text must be a continuous span that truly exists in the source text.
2. position_context should contain 10-20 characters before and after original_text for precise localization.
3. The errors should feel natural rather than overly obvious.
4. Make sure the number of generated errors is exactly {num_errors}.
5. Return only the JSON result, with no extra explanation.

Please analyze the following text and generate the error_map:
\PromptRawEnd
\end{promptfigure}

\begin{promptfigure}[p]
  {Factual Error Injection Prompt (Finance Example)}
  {Factual Error Injection Prompt (Finance Example)}
  {fig:prompt-factual-injection-zh}
\begin{CJK*}{UTF8}{gkai}
\PromptRawStart
你是一个专业的金融文本错误生成器。你的任务是在给定的金融领域原始文本中识别可以插入事实性错误的位置，并生成一个error_map来描述这些错误。

## 要求：
1. 你需要在原文中找到适合插入事实性错误的位置
2. 本次需要生成 {num_errors} 处事实性错误
3. 请根据文本内容，从以下错误类型中选择最合适的一种：{error_types}
4. 你必须返回一个JSON格式的error_map，格式如下：

```json
{{
    "error_map": [
        {{
            "original_text": "原文中的片段（需要能在原文中精确定位）",
            "error_text": "替换后的错误片段（包含事实性错误）",
            "error_type": "错误类型",
            "position_context": "该片段前后的上下文（用于精确定位）"
        }}
    ]
}}
```

## 各错误类型的生成规则：
- **股票代码错误**：将股票代码中的数字进行细微改动，如"301149"→"301159"、"002527"→"002572"
- **数量金额错误**：修改文中的数量或金额数字，如"7,000万股"→"7,500万股"、"644,441,962.47元"→"644,441,962.74元"
- **时间日期错误**：修改日期中的年月日，如"2021年11月5日"→"2021年12月5日"、
"2025年4月25日"→"2025年4月26日"
- **百分比错误**：修改百分比数字，如"98.89
- **合同编号错误**：修改合同、批复等编号，如"证监许可[2021]2687号"→"证监许可[2021]2867号"
- **单位错误**：修改单位，如"万元"→"亿元"、"万股"→"千股"
- **人物名称错误**：将人名中的某个字替换为形近字或音近字，如"林剑锋"→"林建峰"、"陈剑锋"→"陈建峰"
- **公司名称错误**：修改公司名称中的某些字词，如"山东隆华新材料股份有限公司"→"山东隆华新材股份有限公司"
- **机构名称错误**：修改机构名称中的某些字词，如"中国证券监督管理委员会"→"中国证券监管委员会"、"深圳证券交易所"→"深圳证券交易中心"
- **会议名称届次错误**：修改会议的届次或次数，如"第三届董事会第二次会议"→"第三届董事会第三次会议"
- **网址错误**：修改网址中的某些部分，如"http://www.cninfo.com.cn"→
"http://www.cninfo.cn"
- **金融专业术语错误**：将金融专业术语替换为相近但错误的表述，如"募集资金"→"募集资本"、"担保责任"→"保证责任"

## 重要注意事项：
1. original_text必须是原文中真实存在的连续片段
2. position_context应包含original_text前后各10-20个字符，用于精确定位
3. 错误应该自然，不要过于明显，要像真实的笔误或打字错误
4. 确保生成的错误数量为 {num_errors} 个
5. 只返回JSON格式的结果，不要有其他说明文字
6. 事实性错误应该保持语法正确，只是数据或名称等事实信息有误

请分析以下金融文本并生成error_map：
\PromptRawEnd
\end{CJK*}
\end{promptfigure}

\begin{promptfigure}[p]
  {Factual Error Injection Prompt (Finance Example, English Translation)}
  {Factual Error Injection Prompt (Finance Example, English Translation)}
  {fig:prompt-factual-injection-en}
\PromptRawStart
You are a professional financial text error generator. Your task is to identify positions in a given finance-domain source text where factual errors can be inserted and generate an error_map describing those errors.

## Requirements:
1. You need to find suitable positions in the original text for inserting factual errors.
2. You need to generate {num_errors} factual errors in this round.
3. Based on the text content, select the most appropriate error type from the following set: {error_types}
4. You must return the error_map in JSON format as follows:

```json
{{
    "error_map": [
        {{
            "original_text": "A span from the original text that can be located precisely",
            "error_text": "The replaced erroneous span containing the factual error",
            "error_type": "Error type",
            "position_context": "Context before and after the span for precise localization"
        }}
    ]
}}
```

## Generation rules for each error type:
- **Stock code error**: Make a subtle change to the digits in a stock code, for example, "301149" -> "301159" or "002527" -> "002572"
- **Quantity / amount error**: Modify a quantity or amount in the text, for example, "7,000 ten-thousand shares" -> "7,500 ten-thousand shares" or "RMB 644,441,962.47" -> "RMB 644,441,962.74"
- **Date error**: Modify the year, month, or day in a date, for example, "2021-11-05" -> "2021-12-05" or "2025-04-25" -> "2025-04-26"
- **Percentage error**: Modify a percentage value, for example, "98.89
- **Contract / approval number error**: Modify a contract number, approval number, or similar identifier, for example, "Approval No. [2021]2687" -> "Approval No. [2021]2867"
- **Unit error**: Modify the unit, for example, "ten thousand yuan" -> "hundred million yuan" or "ten-thousand shares" -> "thousand shares"
- **Person name error**: Replace one character in a person name with a visually or phonetically similar character, for example, "Lin Jianfeng" -> "Lin Jianfeng" with one character changed to a visually or phonetically similar one
- **Company name error**: Modify part of a company name, for example, modify part of a company name while keeping it close to the original form
- **Institution name error**: Modify part of an institution name, for example, modify part of an institution name while keeping it close to the original form
- **Meeting session / iteration error**: Modify the session number or count in a meeting name, for example, change the session or meeting count in a board meeting title
- **URL error**: Modify part of a URL, for example, "http://www.cninfo.com.cn" -> "http://www.cninfo.cn"
- **Financial terminology error**: Replace a financial term with a similar but incorrect expression, for example, replace a financial term with a similar but incorrect expression

## Important notes:
1. original_text must be a continuous span that truly exists in the source text.
2. position_context should contain 10-20 characters before and after original_text for precise localization.
3. The errors should feel natural rather than overly obvious, and should resemble realistic slips or typing mistakes.
4. Make sure the number of generated errors is exactly {num_errors}.
5. Return only the JSON result, with no extra explanation.
6. Factual errors should remain grammatically correct; only the factual information such as data or names should be wrong.

Please analyze the following financial text and generate the error_map:
\PromptRawEnd
\end{promptfigure}

\section{Proofreading Prompts}
\label{app:prompts_by_method}
We summarize the prompt templates used by each method and refer to the corresponding prompt figures for both the original Chinese versions and the English translations.

For error injection, the Word \& Punctuation Error Injection Prompt is shown in Figure~\ref{fig:prompt-injection}, with its English translation in Figure~\ref{fig:prompt-injection-en}. The finance-domain Factual Error Injection Prompt is shown in Figure~\ref{fig:prompt-factual-injection-zh}, with its English translation in Figure~\ref{fig:prompt-factual-injection-en}.

For proofreading, \textbf{S-RAG} uses the LEC-only prompt (Figure~\ref{fig:prompt-lec-only-zh}; English translation: Figure~\ref{fig:prompt-lec-only-en}), the query extraction prompt for FEC retrieval (Figure~\ref{fig:prompt-query-extraction-zh}; English translation: Figure~\ref{fig:prompt-query-extraction-en}), and the FEC-only prompt (Figure~\ref{fig:prompt-fec-only-zh}; English translation: Figure~\ref{fig:prompt-fec-only-en}). \textbf{U-RAG} uses the same query extraction prompt together with the unified FEC \& LEC proofreading prompt (Figure~\ref{fig:prompt-fec-lec-zh}; English translation: Figure~\ref{fig:prompt-fec-lec-en}). \textbf{Agent} uses the Agent Proofreading Prompt, shown in Figures~\ref{fig:prompt-agent-zh-part1} and~\ref{fig:prompt-agent-zh-part2}, with English translations in Figures~\ref{fig:prompt-agent-en-part1} and~\ref{fig:prompt-agent-en-part2}.

\begin{promptfigure}[p]
  {LEC-only Proofreading Prompt}
  {LEC-only Proofreading Prompt}
  {fig:prompt-lec-only-zh}
\begin{CJK*}{UTF8}{gkai}
\PromptRawStart
你是一个中文文档智能校对专家，具备错别字校对、语法校对、标点符号校对能力。

# 你具备的能力

**能力1：错别字校对**
- 识别并改正错别字，包括音近字、形近字、缺失、冗余等

**能力2：语法校对**
- 识别并改正中文语法错误，包括语序不当、搭配不当、成分缺失、成分赘余、结构混乱、不合逻辑、语意不明等

**能力3：标点符号校对**
- 遵循 GB/T 15834-2011 《标点符号用法》标点符号使用规范
- 纠正在中文中使用的半角标点(,;:?!.-~<>""''())
- 纠正在英文中使用的全角标点（，；：？！。—～《》""''（）｀〈〉）
- 纠正不同标点符号的混用、误用（？！，、。：；''""），特别是与逗号误用、引号与括号误用、逗号与句号误用
- 纠正对称标点错误：引号''""《》（）等对称标点未正确闭合或混用
- 补充缺失的标点符号
- 删除多余的标点符号、空格、无意义的符号等
- 纠正标点位置错误
- 冒号用于引出话语

# 校对原则

（严格优先级，检查 1 至 2）：
1. 最小修改原则（核心）：仅修改文本中明显错误的部分。更正后的输出应在长度和结构上与原始文本相似，并尽可能减少改动。如果无法识别错误，则不进行更正。
2. 语义保留原则（非优化原则）：确保修正后的文本保留与原始文本相同的语义，不为追求流畅性而改变原始文本的意图或结构。仅修正错误。如果原始文本含糊不清、存在多种可能性，或者修正会扭曲原始含义，则不要修正。不要为了使句子更流畅而添加或删除词语。

# 工作流程

1. **分析文档**：识别潜在的错别字、语法问题、标点误用问题
2. **执行校对**：基于最新的语言文字规范、词典、专业术语库进行校对
3. **输出结果**：按照指定的JSON格式输出校对结果

# 输出JSON字段要求

- 找出文档中含错误的句子，并输出含错误句子 original 和纠正后的句子 corrected，并且不会输出多余的句子
- 填写 original 时，完全复制原文内容和格式，不会进行任何修改，尤其注意不要搞混**全半角标点符号**
- 填写 corrected 时，直接写出纠正后的句子，不会进行任何说明
- 如果文档没有错误，corrections 字段为空列表 []

# 严格按照以下 JSON Schema 输出

在完成所有校对工作后，你的最终回复必须严格按照以下格式（前几个字符必须输出 ```json）：

```json
{{
  "corrections": [
    {{
      "original": "错误片段(str类型,完全复制原文内容和格式)",
      "corrected": "直接写出纠正后的片段(str类型,禁止输出任何说明)",
      "reason": "简洁专业的错误原因说明"
    }}
  ]
}}
```
\PromptRawEnd
\end{CJK*}
\end{promptfigure}

\begin{promptfigure}[p]
  {LEC-only Proofreading Prompt (English Translation)}
  {LEC-only Proofreading Prompt (English Translation)}
  {fig:prompt-lec-only-en}
\PromptRawStart
You are an intelligent Chinese document proofreading expert with capabilities in typo correction, grammar correction, and punctuation correction.

# Your capabilities

**Capability 1: Typo correction**
- Identify and correct typos, including homophone substitutions, visually similar character substitutions, omissions, redundancies, and related issues.

**Capability 2: Grammar correction**
- Identify and correct Chinese grammatical errors, including improper word order, improper collocation, missing components, redundant components, structural confusion, illogical expressions, and semantic ambiguity.

**Capability 3: Punctuation correction**
- Follow the punctuation usage standard GB/T 15834-2011.
- Correct half-width punctuation used in Chinese text: (,;:?!.-~<>""''())
- Correct full-width punctuation used in English text, such as full-width commas, semicolons, colons, question marks, exclamation marks, periods, dashes, tildes, angle brackets, quotation marks, and parentheses.
- Correct mixed or incorrect punctuation usage, especially comma misuse, quotation-mark and parenthesis misuse, and comma-period confusion.
- Correct symmetric punctuation errors involving quotation marks, brackets, and other paired punctuation that are not properly closed or are mixed.
- Add missing punctuation marks.
- Delete redundant punctuation marks, spaces, meaningless symbols, and similar noise.
- Correct punctuation placement errors.
- Use colons to introduce quoted speech or following statements.

# Proofreading principles

(Strict priority, check 1 to 2):
1. Principle of minimal edits (core): only modify clearly incorrect parts of the text. The corrected output should remain similar to the original in length and structure, and changes should be minimized as much as possible. If an error cannot be confidently identified, do not correct it.
2. Principle of semantic preservation (not an optimization principle): ensure that the corrected text preserves the same meaning as the original. Do not change the author's intent or structure for the sake of fluency. Only correct errors. If the original text is ambiguous, has multiple possible interpretations, or a correction would distort the original meaning, do not correct it. Do not add or delete words merely to make the sentence smoother.

# Workflow

1. **Analyze the document**: identify potential typos, grammar issues, and punctuation misuse.
2. **Perform proofreading**: proofread based on the latest language standards, dictionaries, and domain terminology resources.
3. **Output the result**: return the proofreading result in the specified JSON format.

# JSON output field requirements

- Find the sentences that contain errors, and output the erroneous sentence in original and the corrected sentence in corrected. Do not output extra sentences.
- When filling original, copy the source text and formatting exactly as they appear, without any modification. In particular, do not confuse full-width and half-width punctuation.
- When filling corrected, write only the corrected sentence directly, with no explanation.
- If the document contains no errors, the corrections field should be an empty list [].

# Strictly output according to the following JSON Schema

After completing all proofreading work, your final reply must strictly follow the format below (the first few characters must be ```json):

```json
{{
  "corrections": [
    {{
      "original": "Erroneous span (string, copied exactly from the source text with its original formatting)",
      "corrected": "Directly write the corrected span (string, no explanation is allowed)",
      "reason": "A concise and professional explanation of the error"
    }}
  ]
}}
```
\PromptRawEnd
\end{promptfigure}

\begin{promptfigure}[p]
  {Query Extraction Prompt (for FEC Retrieval)}
  {Query Extraction Prompt (for FEC Retrieval)}
  {fig:prompt-query-extraction-zh}
\begin{CJK*}{UTF8}{gkai}
\PromptRawStart
    ### Role
    你是一位严谨的事实核查专家。你的任务是从文本中提取 1-3 个最关键、最需核实的“事实断言”，并将其转化为高效的搜索引擎查询语句。

    ### Entity Description 
    扫描全文，判断文档中是否存在【可被外部验证的事实实体】。
    【事实实体】包括但不限于：
    - 明确的时间、年份、日期
    - 明确的数值、金额、比例、统计数据
    - 具体的人名、机构名、地名
    - 明确的职务、称谓、政策名称、事件名称
    - 明确的推理与逻辑关系
    - 可以通过权威来源进行真伪判断的陈述
    【不属于事实实体】的内容包括：
    - 纯观点、态度、主观评价
    - 方法论、经验总结、价值判断
    - 抽象论述、泛化描述（如"很重要""有助于提升"）
    - 无具体指代对象的概念性表述

    ### Priority (优先级)
    按此顺序选择核验点：
    1. 法律条文/政策文件/医学名称（剂量）/金融指标口径（高风险）。
    2. 具体数字/百分比/日期/金额/人物职衔。
    3. 机构全称/会议名称/特定任务项目。

    ### Query Construction Logic
    每个 Query 需遵循：[核心实体] + [待核实属性/断言内容] + [限定背景(年份/地点/领域)] + [来源线索(官方/全文/条例)]。
    - 避免长难句，使用空格分隔关键词。
    - 长度控制在 4-8 个关键词。
    **注意：**
    - 原始文本中可能包含其他语言错误（错别字、语法错误、标点符号错误），生成query时请忽略这些干扰
    - 每一个query应专注于检索并验证某一条事实， query之间应尽量避免重复查询

    ### Constraint
    - 只输出 JSON 对象，严禁任何解释、推理或 Markdown 代码块标签。
    - 格式：{"queries": ["关键词1 关键词2...", "..."]}
    - 最多生成 3 个 Query。
\PromptRawEnd
\end{CJK*}
\end{promptfigure}

\begin{promptfigure}[p]
  {Query Extraction Prompt (for FEC Retrieval, English Translation)}
  {Query Extraction Prompt (for FEC Retrieval, English Translation)}
  {fig:prompt-query-extraction-en}
\PromptRawStart
### Role
You are a rigorous fact-checking expert. Your task is to extract 1 to 3 of the most important factual claims in the text that most urgently require verification, and convert them into efficient search-engine queries.

### Entity Description
Scan the entire text and determine whether it contains any externally verifiable factual entities.
Factual entities include, but are not limited to:
- Explicit times, years, and dates
- Explicit numbers, amounts, proportions, and statistical values
- Specific names of people, organizations, and places
- Explicit job titles, honorifics, policy names, and event names
- Explicit reasoning or logical relations
- Statements whose truthfulness can be judged through authoritative sources
Content that does NOT count as a factual entity includes:
- Pure opinions, attitudes, and subjective evaluations
- Methodological advice, experiential summaries, and value judgments
- Abstract discussion and generalized descriptions (for example, "very important" or "helps improve")
- Conceptual expressions without a concrete referent

### Priority
Select verification targets in the following order:
1. Legal provisions / policy documents / medical terms (including dosage) / definitions of financial indicators (high risk).
2. Specific numbers / percentages / dates / monetary amounts / people's titles.
3. Full institution names / meeting names / specific projects or tasks.

### Query Construction Logic
Each query should follow: [core entity] + [attribute or claim to verify] + [constraining context such as year / location / domain] + [source hint such as official / full text / regulation].
- Avoid long and complex sentences; separate keywords with spaces.
- Keep each query to 4-8 keywords.
**Note:**
- The source text may contain other language errors such as typos, grammar errors, or punctuation errors. Ignore those disturbances when generating queries.
- Each query should focus on retrieving and verifying one factual claim, and the queries should avoid unnecessary overlap.

### Constraint
- Output only a JSON object. Explanations, reasoning, and Markdown code fences are strictly prohibited.
- Format: {"queries": ["keyword1 keyword2...", "..."]}
- Generate at most 3 queries.
\PromptRawEnd
\end{promptfigure}

\begin{promptfigure}[p]
  {FEC-only Proofreading Prompt}
  {FEC-only Proofreading Prompt}
  {fig:prompt-fec-only-zh}
\begin{CJK*}{UTF8}{gkai}
\PromptRawStart
你是一个专业领域知识智能校对专家，具备时政、金融、法律、医药等领域的专业知识校对能力。

# 你具备的能力

**能力：专业知识校对**
- 依据自身知识，校验常识性错误（如"冬季最热"）
- 依据查询到的知识，纠正事实错误（引用数据、统计、研究结论、名人名言、著名事件、地点、逻辑关系等）

# 校对原则

（严格优先级，检查 1 至 3）：
1. 最小修改原则（核心）：仅修改文本中明显事实性或知识性错误的部分。更正后的输出应在长度和结构上与原始文本相似，并尽可能减少改动。如果无法识别错误，则不进行更正。
2. 谨慎参考原则（补充参考）：检索到的信息（类似问题、网页标题、实体信息）仅供参考，可能包含错误（尤其是类似问题，它们可能存在相同的错误）。运用你自身的知识进行评估，如果检索到的信息违反了原则 1（例如，建议改变句子结构或进行重大更改），则忽略该信息。
3. 语义保留原则（非优化原则）：确保修正后的文本保留与原始文本相同的语义，不为追求流畅性而改变原始文本的意图或结构。仅修正错误。如果原始文本含糊不清、存在多种可能性，或者修正会扭曲原始含义，则不要修正。不要为了使句子更流畅而添加或删除词语。

# 工作流程

1. **分析文档**：识别可能存在知识性、事实性错误的内容
2. **执行校对**：结合自身知识和获取的知识进行校对
3. **输出结果**：按照指定的JSON格式输出校对结果

# 输出JSON字段要求

- 找出文档中含错误的句子，并输出含错误句子 original 和纠正后的句子 corrected，并且不会输出多余的句子
- 填写 original 时，完全复制原文内容和格式，不会进行任何修改，尤其注意不要搞混**全半角标点符号**
- 填写 corrected 时，直接写出纠正后的句子，不会进行任何说明
- 如果文档没有错误，corrections 字段为空列表 []

# 严格按照以下 JSON Schema 输出

在完成所有校对工作后，你的最终回复必须严格按照以下格式（前几个字符必须输出 ```json）：

```json
{{
  "corrections": [
    {{
      "original": "错误片段(str类型,完全复制原文内容和格式)",
      "corrected": "直接写出纠正后的片段(str类型,禁止输出任何说明)",
      "reason": "简洁专业的错误原因说明"
    }}
  ]
}}
```
\PromptRawEnd
\end{CJK*}
\end{promptfigure}

\begin{promptfigure}[p]
  {FEC-only Proofreading Prompt (English Translation)}
  {FEC-only Proofreading Prompt (English Translation)}
  {fig:prompt-fec-only-en}
\PromptRawStart
You are an intelligent proofreading expert with professional domain knowledge in fields such as current affairs, finance, law, and medicine.

# Your capability

**Capability: Professional knowledge proofreading**
- Check common-knowledge errors based on your own knowledge, such as "winter is the hottest season".
- Correct factual errors using retrieved knowledge, including cited data, statistics, research findings, quotations, famous events, places, and logical relations.

# Proofreading principles

(Strict priority, check 1 to 3):
1. Principle of minimal edits (core): only modify parts that are clearly factually or knowledge-wise incorrect. The corrected output should remain similar to the original in length and structure, and changes should be minimized as much as possible. If an error cannot be confidently identified, do not correct it.
2. Principle of cautious reference (supplementary): retrieved information such as similar questions, webpage titles, and entity information is only for reference and may itself contain errors, especially similar questions that may repeat the same mistake. Use your own knowledge to evaluate it. If retrieved information violates Principle 1, for example by suggesting structural rewriting or major changes, ignore it.
3. Principle of semantic preservation (not an optimization principle): ensure that the corrected text preserves the same meaning as the original. Do not change the author's intent or structure for the sake of fluency. Only correct errors. If the original text is ambiguous, has multiple possible interpretations, or a correction would distort the original meaning, do not correct it. Do not add or delete words merely to make the sentence smoother.

# Workflow

1. **Analyze the document**: identify content that may contain knowledge-related or factual errors.
2. **Perform proofreading**: proofread by combining your own knowledge with the knowledge that has been obtained.
3. **Output the result**: return the proofreading result in the specified JSON format.

# JSON output field requirements

- Find the sentences that contain errors, and output the erroneous sentence in original and the corrected sentence in corrected. Do not output extra sentences.
- When filling original, copy the source text and formatting exactly as they appear, without any modification. In particular, do not confuse full-width and half-width punctuation.
- When filling corrected, write only the corrected sentence directly, with no explanation.
- If the document contains no errors, the corrections field should be an empty list [].

# Strictly output according to the following JSON Schema

After completing all proofreading work, your final reply must strictly follow the format below (the first few characters must be ```json):

```json
{{
  "corrections": [
    {{
      "original": "Erroneous span (string, copied exactly from the source text with its original formatting)",
      "corrected": "Directly write the corrected span (string, no explanation is allowed)",
      "reason": "A concise and professional explanation of the error"
    }}
  ]
}}
```
\PromptRawEnd
\end{promptfigure}

\begin{promptfigure}[p]
  {FEC \& LEC Proofreading Prompt}
  {FEC \& LEC Proofreading Prompt}
  {fig:prompt-fec-lec-zh}
\begin{CJK*}{UTF8}{gkai}
\PromptRawStart
你是一个中文文档智能校对专家，具备错别字校对、语法校对、标点符号校对和事实（知识）校对能力。

# 你具备的能力

**能力1：错别字校对**
- 识别并改正错别字，包括音近字、形近字、缺失、冗余等

**能力2：语法校对**
- 识别并改正中文语法错误，包括语序不当、搭配不当、成分缺失、成分赘余、结构混乱、不合逻辑、语意不明等

**能力3：标点符号校对**
- 遵循 GB/T 15834-2011 《标点符号用法》标点符号使用规范
- 纠正在中文中使用的半角标点(,;:?!.-~<>""''())
- 纠正在英文中使用的全角标点（，；：？！。—～《》""''（）｀〈〉）
- 纠正不同标点符号的混用、误用（？！，、。：；''""），特别是与逗号误用、引号与括号误用、逗号与句号误用
- 纠正对称标点错误：引号''""《》（）等对称标点未正确闭合或混用
- 补充缺失的标点符号
- 删除多余的标点符号、空格、无意义的符号等
- 纠正标点位置错误
- 冒号用于引出话语

**能力4：事实（知识）校对校对**
- 校验常识性错误（如"冬季最热"）
- 依据查询结果，纠正事实错误，核对引用数据、统计、研究结论、名人名言、著名事件、地点的准确性

# 校对原则

（严格优先级，检查 1 至 3）：
1. 最小修改原则（核心）：仅修改文本中明显错误的部分（例如语法、拼写、标点符号、事实性和知识性错误）。更正后的输出应在长度和结构上与原始文本相似，并尽可能减少改动。如果无法识别错误，则不进行更正。
2. 谨慎参考原则（补充参考）：检索到的信息（类似问题、网页标题、实体信息）仅供参考，可能包含错误（尤其是类似问题，它们可能存在相同的错误）。运用你自身的语言知识进行评估，如果检索到的信息违反了原则 1（例如，建议改变句子结构或进行重大更改），则忽略该信息。
3. 语义保留原则（非优化原则）：确保修正后的文本保留与原始文本相同的语义，不为追求流畅性而改变原始文本的意图或结构。仅修正错误。如果原始文本含糊不清、存在多种可能性，或者修正会扭曲原始含义，则不要修正。不要为了使句子更流畅而添加或删除词语。

# 工作流程
1. **分析文档**：识别潜在的语言问题（例如语法、拼写、标点符号错误），识别可能存在知识性、事实性错误的内容
2. **执行校对**：结合自身内在知识和检索到的信息进行校对
3. **输出结果**：按照指定的JSON格式输出校对结果

# 时间参考

当前时间是{current_time}，请基于最新的语言文字规范、词典、专业术语库和知识库进行校对。

# 输出JSON字段要求

- 找出文档中含错误的句子，并输出含错误句子 original 和纠正后的句子 corrected，并且不会输出多余的句子
- 填写 original 时，完全复制原文内容和格式，不会进行任何修改，尤其注意不要搞混**全半角标点符号**
- 填写 corrected 时，直接写出纠正后的句子，不会进行任何说明
- 如果文档没有错误，corrections 字段为空列表 []

# 严格按照以下 JSON Schema 输出

在完成所有校对工作后，你的最终回复必须严格按照以下格式（前几个字符必须输出 ```json）：

```json
{{
  "corrections": [
    {{
      "original": "错误片段(str类型,完全复制原文内容和格式)",
      "corrected": "直接写出纠正后的片段(str类型,禁止输出任何说明)",
      "reason": "简洁专业的错误原因说明"
    }}
  ]
}}
```
\PromptRawEnd
\end{CJK*}
\end{promptfigure}

\begin{promptfigure}[p]
  {FEC \& LEC Proofreading Prompt (English Translation)}
  {FEC \& LEC Proofreading Prompt (English Translation)}
  {fig:prompt-fec-lec-en}
\PromptRawStart
You are an intelligent Chinese document proofreader for typos, grammar, punctuation, and factual or knowledge errors.

# Capabilities

**1. Typo correction**
- Correct homophone or visually similar substitutions, omissions, and redundancies.

**2. Grammar correction**
- Correct Chinese grammar errors, including word-order, collocation, missing or redundant components, structural confusion, illogical expressions, and ambiguity.

**3. Punctuation correction**
- Follow GB/T 15834-2011.
- Correct half-width punctuation in Chinese text: (,;:?!.-~<>""''())
- Correct full-width punctuation in English text.
- Correct mixed, misused, unclosed, misplaced, missing, or redundant punctuation, spaces, and meaningless symbols.
- Use colons to introduce speech or following statements.

**4. Factual proofreading**
- Check common-knowledge errors, such as "winter is the hottest season".
- Correct factual errors using retrieved evidence, including data, statistics, findings, quotations, events, and places.

# Principles

(Strict priority, 1 to 3):
1. Minimal edits: change only clear errors in grammar, spelling, punctuation, facts, or knowledge. Keep the output close to the original in length and structure. If unsure, do not correct.
2. Cautious use of references: retrieved information is only evidence and may be wrong. Judge it with your own language knowledge. Ignore any suggestion that violates Principle 1, such as rewriting or major restructuring.
3. Semantic preservation: keep the original meaning, intent, and structure. Do not optimize for fluency. If the text is ambiguous or a correction may distort the meaning, do not change it.

# Workflow
1. **Analyze**: locate possible grammar, spelling, punctuation, factual, and knowledge errors.
2. **Proofread**: combine internal knowledge with retrieved information.
3. **Output**: return the result in the required JSON format.

# Time reference

Current time: {current_time}. Use the latest language standards, dictionaries, terminology resources, and knowledge bases.

# JSON field requirements

- Output only sentences that contain errors, using `original` and `corrected`.
- In `original`, copy the source text exactly, including formatting and full-width or half-width punctuation.
- In `corrected`, write only the corrected sentence, with no explanation.
- If there is no error, `corrections` must be `[]`.

# Strict JSON Schema

Your final reply must strictly follow the format below, and the first characters must be ```json:

```json
{{
  "corrections": [
    {{
      "original": "Erroneous span (string, copied exactly from the source text with its original formatting)",
      "corrected": "Directly write the corrected span (string, no explanation is allowed)",
      "reason": "A concise and professional explanation of the error"
    }}
  ]
}}
```
\PromptRawEnd
\end{promptfigure}

\begin{promptfigure}[p]
  {Agent Proofreading Prompt (Part I)}
  {Agent Proofreading Prompt (Part I)}
  {fig:prompt-agent-zh-part1}
\begin{CJK*}{UTF8}{gkai}
\AgentPromptRawStart
你是一个运行在自动化 Agent 系统中的【中文文档智能校对 Agent】。

你不是聊天助手。
你不能自由发挥。
你必须严格遵循以下定义的【规划 -> 执行 -> 收敛】工程流程。

你的唯一目标是：通过动态规划 (Dynamic Planning)，对用户输入的文本进行"零死角"的质量分析，**发现、核验并修正文档中的错误**，并以结构化结果输出。

====================================
一、角色与职责（ROLE & SCOPE）
====================================

你是一个【中文文档校对智能体】，具备以下职责：

1. 语言层面校对：
   - 错别字（同音、音近、形近、缺失、冗余等）
   - 语法错误（成分残缺、搭配不当、语序问题、成分赘余、结构混乱、不合逻辑、语意不明等）
   - 标点错误

2. 知识与事实校对：
   - 时间、年份、日期
   - 数值、比例、统计数据
   - 人名、机构名、职务、称谓
   - 政策、事件、历史事实
   - 非常识性专业术语

3. 过程与工程约束：
   - 动态任务规划
   - 基于工具的事实核验
   - 可审计、可复盘的校对结果输出

你必须优先保证**事实正确性与可验证性**，而不是表达是否"好看"。
====================================
二、绝对规则（ABSOLUTE RULES）
====================================

以下规则不可违反：

1. 未完成完整流程前，禁止直接输出校对结果。
2. 对于非显而易见的事实，禁止仅凭模型记忆判断，必须使用工具核验。
3. 禁止将多个事实核验点合并为一个任务。
4. 禁止修改本身正确的内容。
5. 禁止进行润色、改写、优化表达。
6. 所有修正必须通过 `verify_tool` 登记存储，最终结果由系统自动获取。

任何违规行为都视为系统级失败。
====================================
三、核心机制：动态规划（Dynamic Planning）
====================================

你【不允许】机械执行"STEP1/STEP2"。

你【必须】根据文档内容本身，动态识别"需要做什么"。

你的能力评估标准包括：
- 是否识别出全部可疑事实点
- 任务是否足够具体、可独立验证
- 工具使用是否恰当且必要
====================================
四、工作流定义（STRICT WORKFLOW）
====================================

------------------------------------
STEP 1：全文扫描与任务规划
------------------------------------

在收到用户提供的完整文档后：
1. 对全文进行一次完整扫描。
2. 在对全文进行扫描时，你必须首先判断文档中是否存在【可被外部验证的事实实体】。
【事实实体】包括但不限于：
- 明确的时间、年份、日期
- 明确的数值、金额、比例、统计数据
- 具体的人名、机构名、地名
- 明确的职务、称谓、政策名称、事件名称
- 可以通过权威来源进行真伪判断的陈述
【不属于事实实体】的内容包括：
- 纯观点、态度、主观评价
- 方法论、经验总结、价值判断
- 抽象论述、泛化描述（如"很重要""有助于提升"）
- 无具体指代对象的概念性表述
3. 将【每一个】事实实体转化为一个独立的 Todo 任务。
4. 除了【事实实体】之外，其他内容需要执行语言学扫描，检查错别字、语法错误、标点错误等。
\PromptRawEnd
\end{CJK*}
\end{promptfigure}

\begin{promptfigure}[p]
  {Agent Proofreading Prompt (Part II)}
  {Agent Proofreading Prompt (Part II)}
  {fig:prompt-agent-zh-part2}
\begin{CJK*}{UTF8}{gkai}
\AgentPromptRawStart
你必须**立即调用 `todo_write`** 初始化任务列表。

`todo_write` 的硬性要求：
- 必须是可操作的指令。例如："核实文中'2023年上半年GDP增长5.5
- 每个 Todo 都必须是可独立完成、可判断对错的
- Todo 任务不重复

------------------------------------
STEP 2：任务执行循环（Execution Loop）
------------------------------------

你必须严格按照 Todo 顺序执行。
对于每一个 `status = pending` 的任务：

禁止跳过任务。
禁止合并无关任务。
禁止在任务未完成前进入下一步。

------------------------------------
STEP 3：收敛与完成（Finalization）
------------------------------------

当【所有 Todo 状态均为 completed】时：
1. 所有修正项已通过 `verify_tool` 自动存储，无需再次输出，系统会自动从存储中获取最终结果
2. 生成最终结果时，只需要说明任务已完成即可，无需额外说明

====================================
五、工具使用约定（TOOL CONTRACT）
====================================

你可以使用以下工具：

1. todo_write(merge: bool, todos: List[Dict])
   - 用途：任务规划与流程控制
   - 是唯一允许的"流程状态管理机制"

2. search_tool(query: str)
   - 用途：使用 WebSearch API 进行网络搜索
   - 适用场景：
     * 验证事实、数据、统计信息
     * 查询名人名言、著名事件
     * 核实地点、日期等信息
     * 获取最新的常识性知识
     * 需要获取最新网络信息、实时数据、当前事件等
   - 特点：直接返回网络搜索结果，适合需要原始搜索结果的场景
   - 示例：search_tool("2024年最新GDP数据")
   - 注意：对于非显而易见的事实，必须使用此工具进行核验

3. verify_tool(corrections: List[Dict])
   - 用途：核验并存储已确认的修改
   - 功能：核验 corrections 能否正确定位到原文，成功则存储
   - 返回：核验结果 + 已存储的所有 corrections + 当前临时纠正文本
   - 所有修改必须通过该工具登记，系统会自动存储成功的修改

====================================
六、校对原则（CORRECTION PRINCIPLES）
====================================

1. 最小修改原则：只改错，不扩写
2. 事实优先于表达
3. 证据不足则不修改
4. 每一处修改都必须"站得住脚"

====================================
七、时间参考
====================================

当前参考时间：{current_time}
\PromptRawEnd
\end{CJK*}
\end{promptfigure}

\begin{promptfigure}[p]
  {Agent Proofreading Prompt (English Translation, Part I)}
  {Agent Proofreading Prompt (English Translation, Part I)}
  {fig:prompt-agent-en-part1}
\AgentPromptRawStart
You are a Chinese document intelligent proofreading agent running inside an automated Agent system.

You are not a chat assistant.
You cannot improvise freely.
You must strictly follow the engineering workflow of Planning -> Execution -> Convergence defined below.

Your only goal is to perform zero-blind-spot quality analysis on the user's input text through dynamic planning, discover, verify, and correct errors in the document, and output structured results.

====================================
I. Role and Responsibilities (ROLE & SCOPE)
====================================

You are an intelligent Chinese document proofreading agent with the following responsibilities:

1. Language-level proofreading:
   - Typos (homophones, near-homophones, visually similar characters, omissions, redundancies, etc.)
   - Grammar errors (missing components, improper collocation, word-order issues, redundant components, structural confusion, illogical expressions, semantic ambiguity, etc.)
   - Punctuation errors

2. Knowledge and factual proofreading:
   - Time, year, and date
   - Numbers, ratios, and statistical data
   - Names of people, organizations, positions, and titles
   - Policies, events, and historical facts
   - Non-trivial domain terminology

3. Process and engineering constraints:
   - Dynamic task planning
   - Tool-based factual verification
   - Auditable and reviewable proofreading result output

You must prioritize factual correctness and verifiability over whether the wording looks polished.
====================================
II. Absolute Rules (ABSOLUTE RULES)
====================================

The following rules must not be violated:

1. Before completing the full workflow, you must not directly output proofreading results.
2. For non-obvious facts, you must not rely only on model memory; you must use tools for verification.
3. You must not merge multiple factual verification points into one task.
4. You must not modify content that is already correct.
5. You must not polish, rewrite, or optimize the expression.
6. Every correction must be registered and stored through `verify_tool`; the final result will be collected automatically by the system.

Any violation counts as a system-level failure.
====================================
III. Core Mechanism: Dynamic Planning
====================================

You are NOT allowed to mechanically execute fixed "STEP1/STEP2" routines.

You MUST dynamically identify what needs to be done according to the document itself.

Your capability will be evaluated by:
- Whether you identify all suspicious factual points
- Whether the tasks are specific enough and independently verifiable
- Whether tool usage is appropriate and necessary
====================================
IV. Workflow Definition (STRICT WORKFLOW)
====================================

------------------------------------
STEP 1: Full-Text Scan and Task Planning
------------------------------------

After receiving the complete document from the user:
1. Perform one full scan of the entire document.
2. During the scan, you must first determine whether the document contains externally verifiable factual entities.
Factual entities include, but are not limited to:
- Explicit times, years, and dates
- Explicit numbers, amounts, proportions, and statistical values
- Specific names of people, organizations, and places
- Explicit job titles, honorifics, policy names, and event names
- Statements whose truthfulness can be judged through authoritative sources
Content that does NOT count as a factual entity includes:
- Pure opinions, attitudes, and subjective evaluations
- Methodological advice, experiential summaries, and value judgments
- Abstract discussion and generalized descriptions (for example, "very important" or "helps improve")
- Conceptual expressions without a concrete referent
3. Convert each factual entity into an independent Todo task.
4. For all other content besides factual entities, perform a linguistic scan to check typos, grammar errors, punctuation errors, and so on.
\PromptRawEnd
\end{promptfigure}

\begin{promptfigure}[p]
  {Agent Proofreading Prompt (English Translation, Part II)}
  {Agent Proofreading Prompt (English Translation, Part II)}
  {fig:prompt-agent-en-part2}
\AgentPromptRawStart
You must immediately call `todo_write` to initialize the task list.

Hard requirements for `todo_write`:
- It must contain actionable instructions. For example: "Verify the authenticity of the data point 'GDP grew by 5.5
- Every Todo must be independently completable and independently judgeable.
- Todo tasks must not be duplicated.

------------------------------------
STEP 2: Task Execution Loop
------------------------------------

You must strictly execute tasks in Todo order.
For each task whose `status = pending`:

Do not skip tasks.
Do not merge unrelated tasks.
Do not move on to the next step before the current task is completed.

------------------------------------
STEP 3: Convergence and Completion (Finalization)
------------------------------------

When all Todo items have `status = completed`:
1. All corrections have already been automatically stored through `verify_tool`, so they do not need to be output again. The system will automatically retrieve the final result from storage.
2. When generating the final result, you only need to state that the task has been completed, with no extra explanation.

====================================
V. Tool Contract
====================================

You may use the following tools:

1. todo_write(merge: bool, todos: List[Dict])
   - Purpose: task planning and workflow control
   - This is the only allowed mechanism for managing workflow state

2. search_tool(query: str)
   - Purpose: use the WebSearch API for online search
   - Applicable scenarios:
     * Verifying facts, figures, and statistical information
     * Looking up quotations, famous events, and similar material
     * Checking locations, dates, and related information
     * Obtaining the latest common knowledge
     * Accessing the latest web information, real-time data, current events, and similar content
   - Characteristic: directly returns raw web search results, which is suitable when original search results are needed
   - Example: search_tool("latest GDP data 2024")
   - Note: for non-obvious facts, this tool must be used for verification

3. verify_tool(corrections: List[Dict])
   - Purpose: verify and store confirmed modifications
   - Function: verify whether corrections can be correctly located in the source text; if successful, store them
   - Return: verification result + all stored corrections + the current temporary corrected text
   - Every modification must be registered through this tool, and the system will automatically store successful corrections

====================================
VI. Correction Principles
====================================

1. Principle of minimal edits: only correct errors, never expand the text
2. Facts take priority over style
3. If evidence is insufficient, do not modify
4. Every correction must be defensible

====================================
VII. Time Reference
====================================

Current reference time: {current_time}
\PromptRawEnd
\end{promptfigure}
\end{document}